\documentclass[journal]{IEEEtran}

\usepackage{times}  
\usepackage{helvet} 
\usepackage{courier}  
\usepackage[hyphens]{url}  
\usepackage{graphicx} 
\usepackage{graphicx}  
\usepackage{color}
\usepackage{caption}
\usepackage{cite}

\usepackage{caption} 

\usepackage{epsfig}
\usepackage{amsmath}
\usepackage{amssymb}

\usepackage[hidelinks]{hyperref}
\usepackage{color}
\usepackage{threeparttable}
\usepackage{longtable}
\usepackage{rotating}
\usepackage{tabularx}
\usepackage{enumitem}
\usepackage{bbm}
\usepackage{pifont}
\usepackage{amssymb}
\usepackage{arydshln}
\usepackage{indentfirst}  

\usepackage{amsfonts}

\usepackage{booktabs}
\usepackage{color}
\usepackage[noend]{algpseudocode}

\usepackage{algorithmicx,algorithm}

\usepackage[switch]{lineno}
\usepackage{soul, color, xcolor}

\soulregister\cite7 
\soulregister\citep7 
\soulregister\citet7 
\soulregister\ref7 
\soulregister\ding7 
\soulregister\pageref7 

\usepackage{booktabs}
\usepackage{multirow}
\usepackage{makecell}
\newlength\savewidth
\ifCLASSINFOpdf
\else
\fi
\begin{document}
%
\title{Rethinking the Competition between Detection and ReID in Multi-Object Tracking}
%
%
%

\author{Chao~Liang*,
        Zhipeng~Zhang*,
        Xue~Zhou$\dagger$, 
        Bing~Li,
        Shuyuan~Zhu,
        Weiming~Hu,~\IEEEmembership{Senior Member,~IEEE.}

\thanks{C. Liang and X. Zhou are with School of Automation Engineering, University of Electronic Science and Technology of China (UESTC), Chengdu, China. X. Zhou is also with Shenzhen Institute of Advanced Study, UESTC, Shenzhen, China and Intelligent Terminal Key Laboratory of SiChuan Province. Z. Zhang, B. Li, and W. Hu are with pattern recognition and intelligent system in the National Laboratory of Pattern Recognition (NLPR), Institute of Automation, Chinese Academy of Sciences (CASIA), Beijing, China, and the School of Artificial Intelligence, University of Chinese Academy of Sciences (UCAS), Beijing, China.  W. Hu is also with CAS Center for Excellence in Brain Science and Intelligence Technology, Shanghai, China. S. Zhu is with School of Information and Communication Engineering, University of Electronic Science and Technology of China (UESTC), Chengdu, China. (e-mail: 201921060415@std.uestc.edu.cn, zhangzhipeng2017@ia.ac.cn, zhouxue@uestc.edu.cn)}


}


%
%

\markboth{IEEE TRANSACTIONS ON IMAGE PROCESSING}%
{Shell \MakeLowercase{\textit{et al.}}: Bare Demo of IEEEtran.cls for IEEE Journals}
%



\maketitle

\begin{abstract}
Due to balanced accuracy and speed, one-shot models which jointly learn detection and identification embeddings, have drawn great attention in multi-object tracking (MOT). However, the inherent differences and relations between detection and re-identification (ReID) are unconsciously overlooked because of treating them as two isolated tasks in the one-shot tracking paradigm. This leads to inferior performance compared with existing two-stage methods. In this paper, we first dissect the reasoning process for these two tasks, which reveals that the competition between them inevitably would destroy task-dependent representations learning. To tackle this problem, we propose a novel reciprocal network (REN) with a self-relation and cross-relation design so that to impel each branch to better learn task-dependent representations. The proposed model aims to alleviate the deleterious tasks competition, meanwhile improve the cooperation between detection and ReID. Furthermore, we introduce a scale-aware attention network (SAAN) that prevents semantic level misalignment to improve the association capability of ID embeddings. By integrating the two delicately designed networks into a one-shot online MOT system, we construct a strong MOT tracker, namely CSTrack. Our tracker achieves the state-of-the-art performance on MOT16, MOT17 and MOT20 datasets, without other bells and whistles. Moreover, CSTrack is efficient and runs at 16.4 FPS on a single modern GPU, and its lightweight version even runs at 34.6 FPS. The complete code has been released at \textcolor{magenta}{\url{https://github.com/JudasDie/SOTS}}.

\end{abstract}

\begin{IEEEkeywords}
Multi-object Tracking, Reciprocal Representation Learning, Scale-aware Attention, One-shot, ID embedding.
\end{IEEEkeywords}

\newcommand\blfootnote[1]{%
\begingroup 
\renewcommand\thefootnote{}\footnote{#1}%
\addtocounter{footnote}{-1}%
\endgroup 
}
{
	\blfootnote{*Equal Contribution \quad
 ~$^\dagger$ Corresponding author.
	}
}

%
\IEEEpeerreviewmaketitle

\section{Introduction}
Multi-object tracking (MOT), aiming to estimate the locations and scales of multiple targets in a video sequence, is one of the most fundamental yet challenging tasks in computer vision~\cite{overview}. It may be applied to many practical scenarios, such as intelligent driving, human-computer interaction, and pedestrian behavior analysis.

\begin{figure}[t]
\begin{center}
\includegraphics[width=0.95\linewidth]{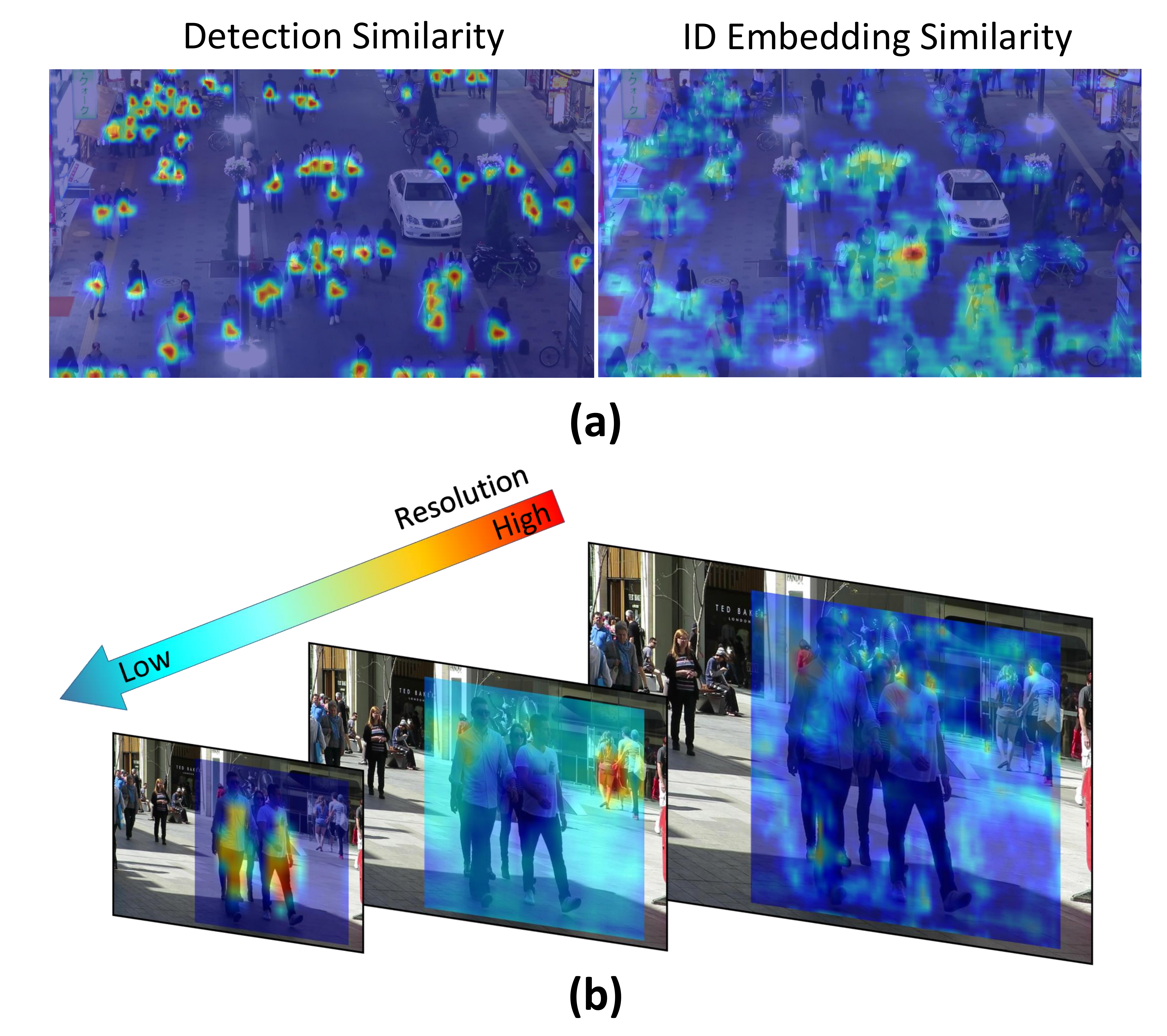}
\end{center}
\vspace{-0.2cm}
\caption{\textbf{Motivation of CSTrack.} (a) visualizes the similarity maps about detection and ReID tasks respectively, where detection expects all the pedestrians have high response values whereas ReID tends to focus on the specific pedestrian. From this point of view, they are contradictory to each other. (b) represents that different resolution focuses on different scale of targets in FPN-based model, where the arrow indicates the output resolution from high to low. This arrangement can help detector to detect pedestrians with different sizes, but not suitable for semantic matching of pedestrians with different sizes in ReID task.}
\label{fig:competition}
\vspace{-0.3cm}
\end{figure}

Convolutional Neural Network (CNN) based object detection and  re-identification (ReID) have demonstrated excellent performance in multi-object tracking in the past few years. The related ReID-based trackers can be divided into two classes, \textit{i.e.,} two-stage and one-shot structures. The two-stage models follow the tracking-by-detection paradigm (or more precisely, tracking-after-detection)~\cite{deepsort,poi,HOGM,RAN}, which divides MOT into two separate tasks, \textit{i.e.,} detection and association. It first obtains bounding boxes of objects in each frame through an off-the-shelf detector, and then associates the candidate boxes with existing tracklets by matching the extracted identification (ID) embedding of each bounding box across frames. Although effective, two-stage methods suffer from massive computational cost, since the ID embedding extraction model (\textit{e.g.}, ReID network~\cite{strongreid}) needs to perform forward inference for each individual bounding box. Alternatively, the one-shot methods~\cite{JDIF,JDE,retinatrack,fairmot} integrate detection and ID embedding extraction into a unified framework. By redesigning the prediction head of the detector, the one-shot tracker can simultaneously yield detection results and ID embeddings, which may bring an increase in speed but at the same time a decrease in accuracy.


In this paper, we first dissect the reasoning process of one-shot tracker and analyze why the performance is lower than the two-stage methods. Our analysis reveals that the performance degradation primarily derives from excessive competition between detection and ReID tasks, which can be summarized into two aspects: \textbf{1) Competition of Object Representation Learning:} In one-shot methods, object class confidence, target scale and ID information are obtained by processing a shared feature tensor. Although one-shot methods are efficient, the inherent differences between different tasks are ignored, as shown in Fig.~\ref{fig:competition} (a). Specifically, object detection requires that objects belonging to the same category (\textit{e.g.}, pedestrian) have similar semantics, while ReID tends to distinguish two different pedestrian, which is contradictory to the purpose of detection task. The competition between the above two tasks may lead to ambiguous learning, which means that the pursuit of high performance in one task may result in performance degradation or stagnation in the other. \textbf{2) Competition of Semantic Level Assignment:} The one-shot trackers~\cite{JDIF,JDE,retinatrack} usually convert detector to tracker by adding a parallel branch to yield ID embeddings. Although it is simple, it ignores that semantic level assignment of detection task is not suitable for ReID task. Specifically, modern object detectors always introduce feature pyramid networks (FPNs)~\cite{FPN} to improve localization accuracy for various target sizes. In this framework, objects of different scales will be assigned to different resolution features, as shown in Fig.~\ref{fig:competition} (b). However, we observe that this assignment is not suitable for ReID in the one-shot tracker since it will lead to a semantic level misalignment, whether for different targets or the same target across frames. Existing one-shot trackers~\cite{JDE,JDIF} usually extract ID embedding from a certain feature resolution depending on detection results, which may lead to a semantic level gap when matching targets with drastic scale changes. Furthermore, it will impair the performance of subsequent data association (Please see the ablation experiments in Sec.~IV-B for detailed discussion and verification).


%

To tackle the above problems, we fully consider the essential differences of two tasks in one-shot tracker, and design two sub-networks to solve the above problems. Specifically, we propose a novel reciprocal network (REN) to conduct collaborative learning among different tasks. The reciprocal network separates the feature maps of object detection and ID embedding extraction into two different task-driven branches, by which the task-dependent representations can be learned. Specifically, given shared features, a novel structure combining self-relation and cross-relation is designed to enhance the feature representation. The former impels hidden nodes to learn task-dependent features, and in parallel the latter aims to improve the collaborative learning of these two tasks. Meanwhile, we design a scale-aware attention network (SAAN) to improve the alignment of ID embedding extraction, and enhance the model resilience to scale changes. In this work, we develop an architecture that combines spatial and channel attention to achieve our goal, and this architecture enhances the influence of object-related regions and channels. The output of SAAN is a feature tensor that includes all the targets with rich semantics in all resolutions. Following the above design idea, aggregation feature-based ID embedding is helpful to prevent semantic misalignment during matching.


The main contributions of our work are three-folds: 

\begin{itemize}

 \item We propose a novel reciprocal network to learn task-dependent representations. It not only effectively mitigates the deleterious competition, but also improves the capability of collaborative learning between detection and ReID tasks in one-shot MOT methods.
 
\item We introduce a scale-aware attention network to apply spatial and channel attention to feature maps at different resolutions. By fusing these features and extracting ID embeddings from a consistent representation, it effectively prevents semantic level misalignment, and improves the resilience to objects with different scales during matching.

\item The extensive experiments demonstrate that our method effectively improves the performance of the one-shot MOT method, especially for association ability of re-identification features, which is competitive with the two-stage methods. 
\end{itemize}

The organization of this paper is as follows: Section~\ref{sec:rw} reviews related work in the literature. Section~\ref{sec:method} presents our proposed approach. The experimental results and analysis are provided in Section~\ref{sec:exp}. Section~\ref{sec:conclusion} concludes the paper.

\section{Related Work}
\label{sec:rw}
We firstly review the related MOT methods, including two-stage and one-shot structures. After that, we briefly review the related works for joint detection and tracking without ReID in MOT task.

\subsection{Two-stage MOT Methods}
With the development of object detectors, many MOT trackers follow the tracking-by-detection paradigm (or more precisely, tracking-after-detection). In particular, these trackers~\cite{sort,ioutracker,deepsort,poi,HOGM,RAN} divide MOT into two separate tasks, \textit{i.e.,} detection and association. The object bounding boxes are firstly predicted by high-performance detectors such as Faster R-CNN~\cite{fasterrcnn}, SDP~\cite{SDP} and DPM~\cite{DPM}. Then, the candidate boxes are linked into tracklets across frames by an association network. Since the candidate boxes can be directly yielded by off-the-shelf detector, two-stage MOT methods focus on improving association performance. In the early works, Sort~\cite{sort} firstly performs Kalman filtering to predict candidate boxes in the next frame and uses Hungarian method to associate across frames by measuring bounding box overlap. IOU-Tracker~\cite{ioutracker} directly associates the last frame bounding box with candidate boxes in the current frame using IOU matching. Although simple and efficient, both Sort and IOU-Tracker may fail in challenging scenarios such as incomplete detections or crowded scenes since they rely on the nearest neighbor hypothesis (\textit{i.e.,} only the candidate box closest to the target has the same ID). Therefore, some works~\cite{deepsort,poi,HOGM,RAN} attempt to apply ReID network to yield more robust predictions. In a nutshell, each target is cropped from the image and fed into an additional network to extract ID appearance embedding, which is employed to associate with existing tracklets by measuring the cosine distance across frame. Moreover, some works~\cite{spa,mpntracker} introduce graph neural networks to further improve matching capacity. For example, MPN~\cite{mpntracker} replaces the Hungarian algorithm with graph neural networks to dissect the information and associate them with the edge classification.

Although the above methods achieve impressive performance, they still suffer from massive computation costs since the ID information extraction network needs to perform forward inference for each bounding box. In contrast, one-shot trackers consider constructing an entire model to simultaneously generate detection boxes and ID embeddings, which exhibits attractive efficiency.

\subsection{One-shot MOT Methods}
Due to balanced accuracy and speed, the one-shot paradigm~\cite{JDIF,JDE,retinatrack,fairmot}, which integrates detection and ID embedding extraction into a unified network, offer a new solution for MOT. Tong \textit{et al.}~\cite{JDIF} first modify a detector (\textit{i.e.,} Faster RCNN~\cite{fasterrcnn}) to handle both detection and ReID tasks . By introducing extra fully-connected layers (FC), the detector obtains the ability to yield ID embeddings. Recently-proposed JDE~\cite{JDE} and RetinaTrack~\cite{retinatrack} convert a FPN-based detector ( \textit{i.e.,} Yolo-v3~\cite{yolov3} and RetinaNet~\cite{retinanet}) to a one-shot tracker by redesigning the prediction head. Although those simple modifications help generate detection and ID embeddings with a small overhead and achieve impressive performance, they ignore the inherent differences between detection and ReID. This induces their performance not so good as the two-stage methods, especially evaluated by the IDF1 score. Recent released method, FairMOT~\cite{fairmot},  uses the anchor-free method~\cite{centernet} to reduce the ambiguity of anchors. Despite alleviating the scale-aware competition by aggregating multi-layer features to yield ID embeddings, it fails to handle the competition between object detection and ReID. 


\subsection{Joint Detection and Tracking without ReID}
Besides the one-shot MOT methods, several methods~\cite{CTrack,tubetk,tracktor,centertrack} attempt to train joint detection and tracking models by integrating other association ways instead of classical ReID. For instance, CTracker~\cite{CTrack} uses adjacent frame pairs as input and generates a pair of bounding boxes for the same target by a chaining structure. TubeTK~\cite{tubetk} utilizes tubes to encode the target's temporal-spatial position and local moving trail. Tracktor~\cite{tracktor} converts existing object detectors (\textit{i.e.,} FasterRCNN~\cite{fasterrcnn}) to trackers by exploiting the regression head of detector to perform the regression of the object boxes from last frame to current image. The following work, CenterTrack~\cite{centertrack}, which shares a similar spirit with Tracktor~\cite{tracktor}, predicts the location of the existing objects by learning the offset of heat points (the center point of object boxes). However, their association performances are also inferior to two-stage methods due to the poor capability of the direct bounding boxes transduction compared with ReID. 



\section{Methodology}
\label{sec:method}
Our proposed framework, named \textbf{CSTrack}, consists of two main components, \textit{i.e.}, reciprocal network designed in Sec.~\ref{sec:REN} and scale-aware attention network described in Sec.~\ref{sec:SAAN}. Before shedding light on the two essential parts, we first give a brief overview of the whole framework in Sec.~\ref{overview}. Moreover, we describe the details of training and online inference in Sec,~\ref{sec:training} and Sec,~\ref{sec:inference}, respectively.

\subsection{Overview} 
\label{overview}

\begin{figure*}[t]
\begin{center}
\includegraphics[width=1\linewidth]{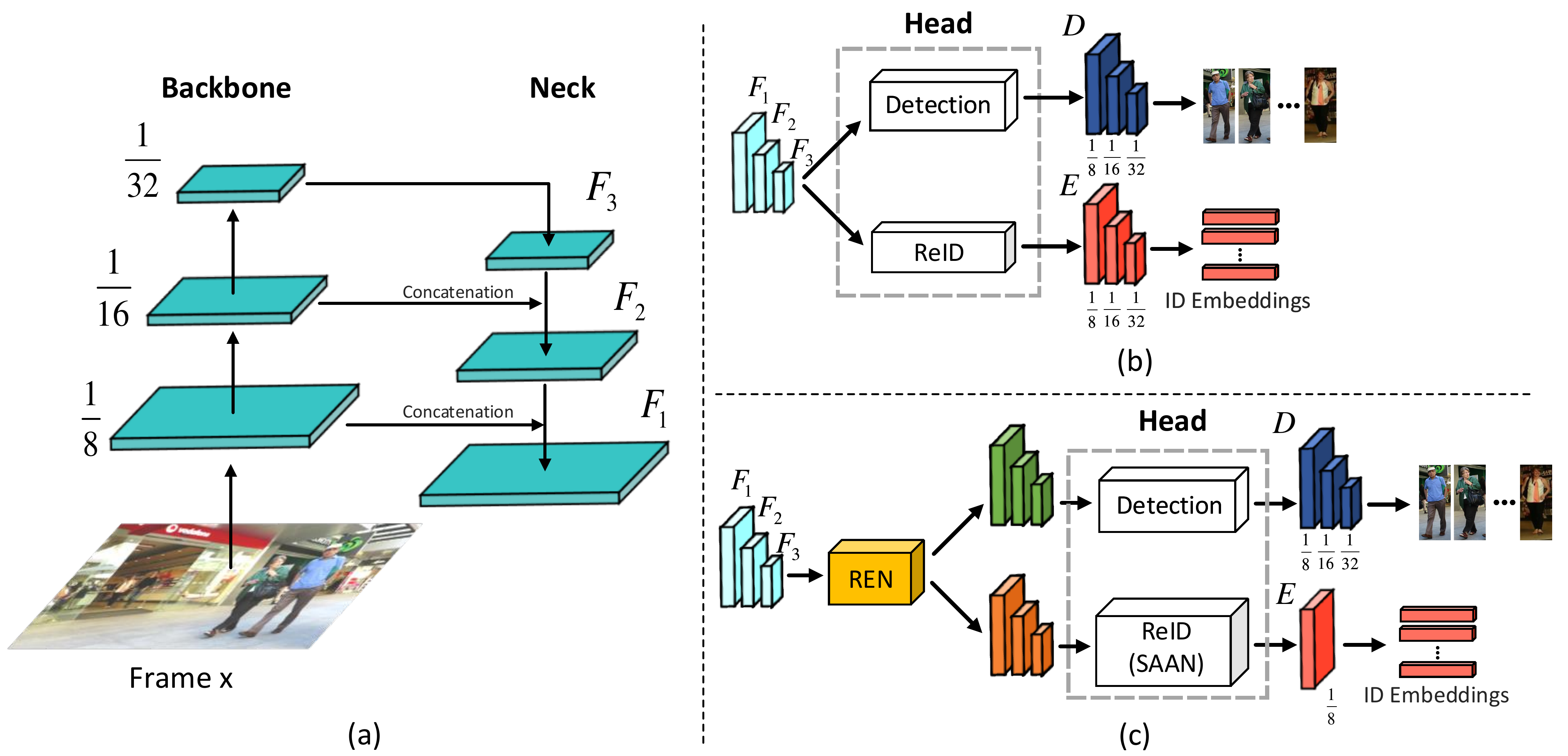}
\end{center}
\caption{Architecture diagrams. (a) is the feature extractor including backbone and neck (FPN). (b) illustrates the vanilla prediction structure of JDE. (c) illustrates our proposed prediction structure of CSTrack.  Different from JDE, CSTrack introduces a reciprocal network (REN) to learn task-dependent representations and a scale-aware attention network to generate discriminative embeddings, which efficiently mitigate the competition.}
\label{fig:overview}
\vspace{-1em}
\end{figure*}

Before describing our model, we review the popular JDE architecture~\cite{JDE} (Joint Detection and Embeddings), which is the baseline of our method, and explain why the vanilla model is not suitable for mitigating the competition between detection and ReID tasks. And then, we introduce the CSTrack framework, which strikes for a better balance between detection and ReID tasks by integrating our proposed sub-networks.


\textbf{Preliminaries.} JDE~\cite{JDE} redesigns the prediction head of an object detector to conduct object detection and ID embedding extraction simultaneously in a one-shot model. Specifically, given a frame $\boldsymbol{x}$, it is first processed by a feature extractor $\Psi$ (including \textit{Backbone} and \textit{Neck}), which yields multi-resolution features $\boldsymbol{F}_i|_{i=1,2,3}$ (see Fig.~\ref{fig:overview} (a)),
\begin{equation}
\boldsymbol{F}_i|_{i=1,2,3}=\Psi(\boldsymbol{x}).
\end{equation}
Then, $\boldsymbol{F}_i|_{i=1,2,3}$ is fed into the \textit{Head} network (\textit{i.e.}, including the parallel object detection and ReID branches) to predict detection result $\boldsymbol{D}$ and raw ID embedding maps $\boldsymbol{E}$, as illustrated in Fig.~\ref{fig:overview} (b). The detection result will be post-processed by non maximum suppression (NMS)~\cite{fasterrcnn} to yield candidate boxes. Each candidate box extracts its corresponding ID embedding from the row ID embedding maps $\boldsymbol{E}$ (following the same feature resolution level as the candidate box) to link with the existing tracklets.

In this framework, the shared features are directly fed into two independent task-driven branches, which generate the output directly applying a 1 $\times$ 1 convolution layer, ignoring the inherent differences between these two tasks. It may lead to ambiguous learning during training. Moreover, ReID branch extracts ID embeddings from probable arbitrary feature level guided by detection task, which may induce semantic level misalignment during matching. 

\begin{figure*}[t]
\begin{minipage}[b]{1\linewidth}
\includegraphics[width=0.99\linewidth]{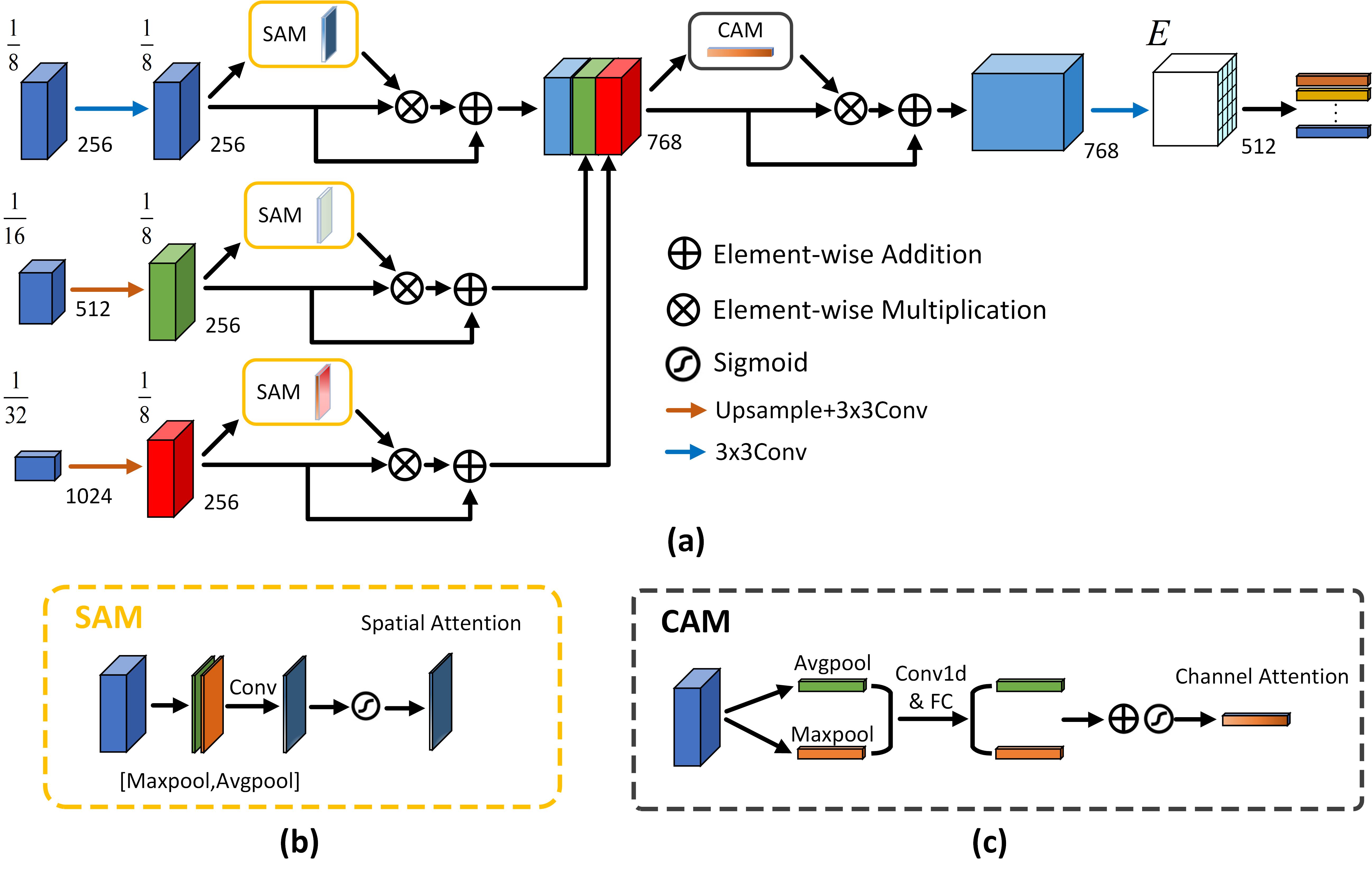}
\end{minipage}
\caption{The details of scale-aware attention network (SAAN). Wherein, (a) is the overall structure of the network, (b) is the diagram of spatial attention module (SAM), and (c) is the diagram of channel attention module (CAM).}
\label{fig:SAAN}
\end{figure*}

\textbf{CSTrack.} We build our framework based on JDE~\cite{JDE}, as shown in Fig.~\ref{fig:overview} (c). To alleviate the competition of object representation learning and enhance task-dependent representations, we propose a novel reciprocal network (REN) to decouple the feature $\boldsymbol{F}_t$ before feeding it into different task branches and then exchange semantic information of different tasks with a cross-relation layer. The idea of designing REN is inspired by recent self-attention~\cite{selfattention} and multi-task decoding mechanisms~\cite{padnet}, which can enhance representations for each task with only small overheads. Note that, REN not only focuses on the specificities of features for different tasks, but also learns commonalities to encode the feature by constructing the cross-relation weight maps. For ReID branch, instead of applying a $1 \times 1$ convolution layer on each feature resolution, we design a scale-aware attention network (SAAN, detailed in Fig.~\ref{fig:SAAN}) to fuse features from different feature resolutions. Meanwhile, both spatial and channel-wise attention modules~\cite{cbam} are adopted to suppress noisy background and learn object-related representations. 


\subsection{Reciprocal Network}
\label{sec:REN}

\begin{figure}[t]
\begin{minipage}[b]{1\linewidth}
\centerline{\includegraphics[height=12\baselineskip]{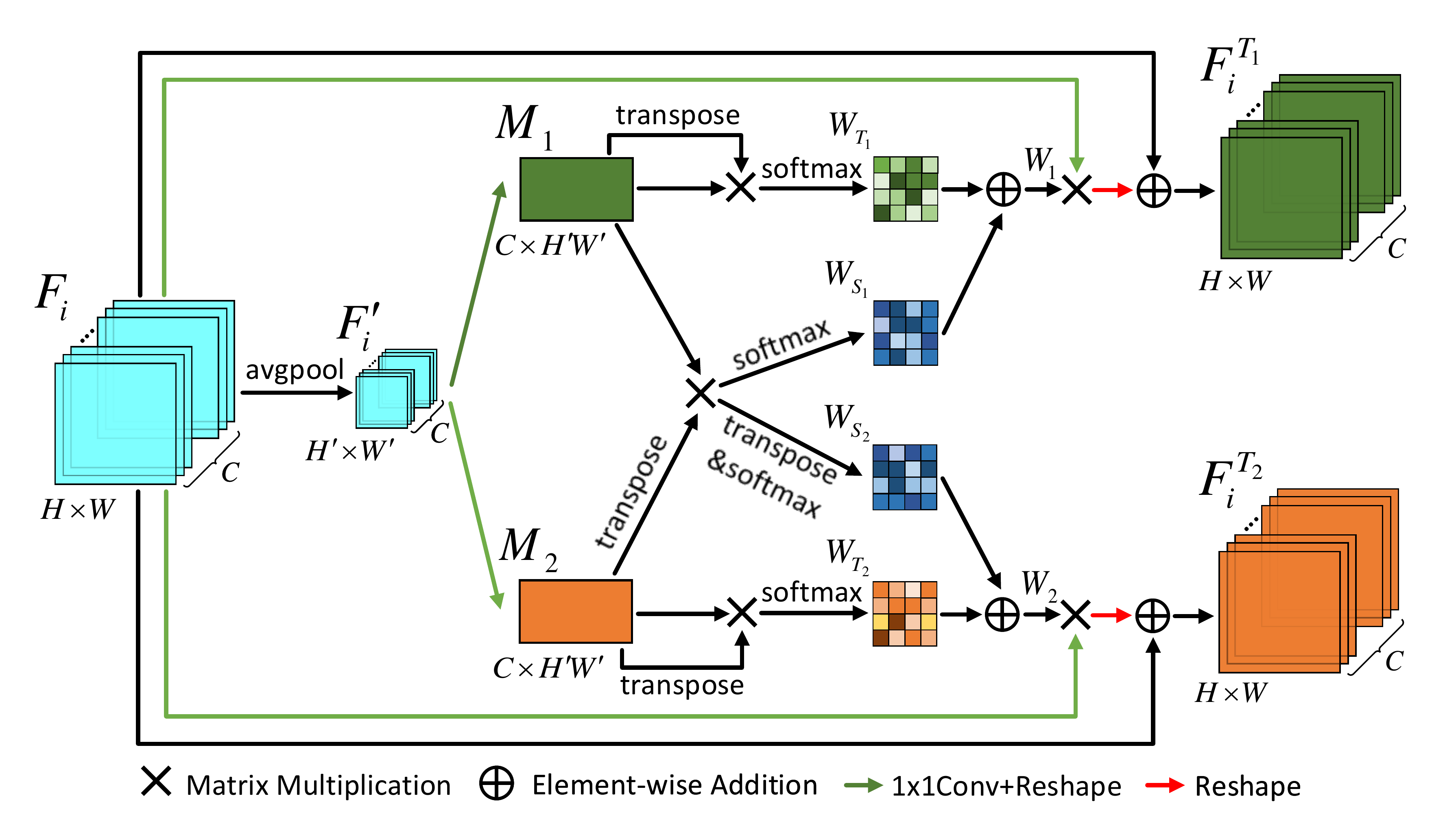}}
\caption{Diagram of reciprocal network (REN). For the original feature map $\boldsymbol{F}_i$, we construct the self-relation and cross-relation maps to impel the generation of task-dependent features $\boldsymbol{F}_i^{T_1}$ and $\boldsymbol{F}_i^{T_2}$.}
\label{REN}
\end{minipage}
\end{figure}

In this section, we propose a reciprocal network to learn both specificities (task-dependent) and commonalities (collaboration) of features for detection and ReID tasks. For specificities learning, self-relation reflecting correlations between different feature channels are learned to enhance feature representation for each task. For commonalities learning, a cross-relation layer is elaborately designed to exchange the semantic information of different tasks.

The structure of our reciprocal network is illustrated in Fig. \ref{REN}, where the features outputted from feature extractor are denoted as $\boldsymbol{F}_i|_{i=1,2,3} \in \mathbb{R}^{C \times H \times W}$. First, we pass $\boldsymbol{F}_i$ through an \textit{avg-pooling} layer to obtain the statistical information $\boldsymbol{F}_i^{\prime} \in \mathbb{R}^{C \times H^{\prime} \times W^{\prime}}$. Second, tensors $\boldsymbol{M}_{1}$ and $\boldsymbol{M}_{2}$ for detection and ReID are respectively generated by passing $\boldsymbol{F}_i^{\prime}$ through different convolution layers and reshape them to $C \times N^{\prime}$, where $N^{\prime}=H^{\prime} \times W^{\prime}$. Third, we perform matrix multiplication on $\boldsymbol{M}_{1}/\boldsymbol{M}_{2}$ ($\mathbf{/}$ indicates ``or'') and its corresponding transpose tensor. A row \textit{softmax} layer is followed to calculate the self-relation weight maps $\left\{\boldsymbol{W}_{T_{1}}, \boldsymbol{W}_{T_{2}}\right\} \in \mathbb{R}^{\mathrm{C \times C}}$ for each task, and the calculation is as follows:
{\setlength\abovedisplayskip{5pt}
\setlength\belowdisplayskip{5pt}
\begin{equation}
w_{T_k}^{i j}=\frac{\exp \left(\boldsymbol{m}_{k}^{i} \cdot \boldsymbol{m}_{k}^{j}\right)}{\sum_{j=1}^{C} \exp \left(\boldsymbol{m}_{k}^{i} \cdot \boldsymbol{m}_{k}^{j}\right)}, k \in \left\{ 1,2 \right\}
\end{equation}}
where $\cdot$ denotes dot product operation. $\boldsymbol{m}_{k}^{i}$ and $\boldsymbol{m}_{k}^{j}$ indicate the $i^{th}$ and $j^{th}$ row of $\boldsymbol{M}_{1}/\boldsymbol{M}_{2}$, respectively. $w_{T_k}^{i j}$ denotes the value at location of $(i,j)$ on $\boldsymbol{W}_{T_k}$, which represents the relation of the $i^{th}$ and $j^{th}$ channel in a tensor. Then, we perform matrix multiplication between $\boldsymbol{M}_{1/2}$ and the transpose of $\boldsymbol{M}_{2/1}$ to learn commonalities between different tasks, and then a row \textit{softmax} layer is followed to generate cross-relation weight maps $\left\{\boldsymbol{W}_{S_{1}}, \boldsymbol{W}_{S_{2}}\right\} \in \mathbb{R}^{\mathrm{C \times C}}$, 
{\setlength\abovedisplayskip{5pt}
\setlength\belowdisplayskip{5pt}
\begin{equation}
w_{S_k}^{i j}=\frac{\exp \left(\boldsymbol{m}_{k}^{i} \cdot \boldsymbol{m}_{h}^{j}\right)}{\sum_{j=1}^{C} \exp \left(\boldsymbol{m}^{i}_{k} \cdot \boldsymbol{m}_{h}^{j}\right)}, (k,h) \in \left\{ (1,2),(2,1) \right\}
\end{equation}}
where $w_{S_k}^{i j}$ denotes the effect of the $i^{t h}$ feature channel of task 1/2 on the $j^{t h}$ feature channel of task 2/1. Finally, the self-relation and cross-relation weights are finally fused by a  trainable parameter $\lambda$, obtaining $\left\{\boldsymbol{W}_1, \boldsymbol{W}_2\right\} \in \mathbb{R}^{C \times C}$
{\setlength\abovedisplayskip{5pt}
\setlength\belowdisplayskip{-5pt}
\begin{equation}
\boldsymbol{W}_{k} = \lambda_{k} \times \boldsymbol{W}_{T_k} + (1 - \lambda_{k}) \times \boldsymbol{W}_{S_{k}}, k \in \left\{ 1,2 \right\}.
\end{equation}}

The original feature map $\boldsymbol{F}_i$ is processed by one $1 \times 1$ convolution layer and rearranged to the shape of $\mathbb{R}^{C \times N}$, where $N=H \times W$. Then, we perform matrix multiplication between the reshaped feature and the learned weight maps $\boldsymbol{W}_{1/2}$ to obtain an enhanced representation for each task. The enhanced representation is rearranged to the same shape as the original feature map $\boldsymbol{F}_i$ (\textit{i.e.,} $\mathbb{R}^{C \times H \times W}$) and fused with original $\boldsymbol{F}_i$ by residual attention to prevent information loss. The feature tensors $\boldsymbol{F}^{T_i}_i$ and $\boldsymbol{F}^{T_2}_i$ will be sent to different task branch for further processing, respectively.

\subsection{Scale-aware Attention Network}
\label{sec:SAAN}
We build a scale-aware attention network (SAAN), as illustrated in Fig.~\ref{fig:SAAN} to aggregate features from different resolutions, which guarantees semantic level alignment of ID embeddings from objects with different sizes. In this network, we introduce the spacial attention module (SAM)~\cite{cbam} and channel attention module (CAM) to learn `where' and `what' feature is more important for yielding discriminative ID embeddings. In particular, the features from the scale of 1/16 and 1/32 (compared to the size of input image) are upsampled to 1/8 firstly. Then, a $3\times 3$ convolutional layers are followed to encode the upsampled feature maps. For better `awaring' useful information for each target in different resolutions, we firstly arrange SAM to enhance the target-related features and suppress background noise. Specially, we perform \textit{avg-pooling} and \textit{max-pooling} on channel dimension to yield two \textit{2D} maps with size $1 \times H \times W$. These maps are concatenated and processed by a 7 $\times$ 7 convolutional layer and a \textit{Sigmoid} layer sequentially to generate a spatial attention map. The obtained spatial attention map learned individually in each resolution and is fused with the original feature by element-wise multiplication and residual attention. Then, we concatenate the feature maps from different scales and pass them through the channel attention module. The channel attention module comprises global \textit{avg-pooling} and \textit{max-pooling} layers, which learn different statistical information of the different resolution features. The outputs of pooling layers are first processed by the shared network consisting of a \textit{1D convolutional} layer and a \textit{fully-connected} layer. Then, the feature maps will be fused by element-wise addition and normalized by a \textit{Sigmoid} layer to generate an attention map with only one channel, which is applied to the vanilla features similar to the spacial attention map. For the arrangement of the attention module, the experimental result shows that our design is helpful to improve association ability of ID embeddings, as discussed in Sec.~\ref{sec:ablations}.

Finally, we use a $3\times 3$ convolution layer to map features to 512 channels, as $\boldsymbol{E} \in \mathbb{R}^{512 \times W \times H}$. Each ID embedding $\boldsymbol{e_{x y}} \in \mathbb{R}^{512 \times 1 \times 1}$ denotes the identity information of the object at location $(x, y)$, which can be extracted according to detection results for the subsequent ReID task. 

\begin{figure*}[t]
\begin{minipage}[b]{1\linewidth}
\includegraphics[width=1\linewidth]{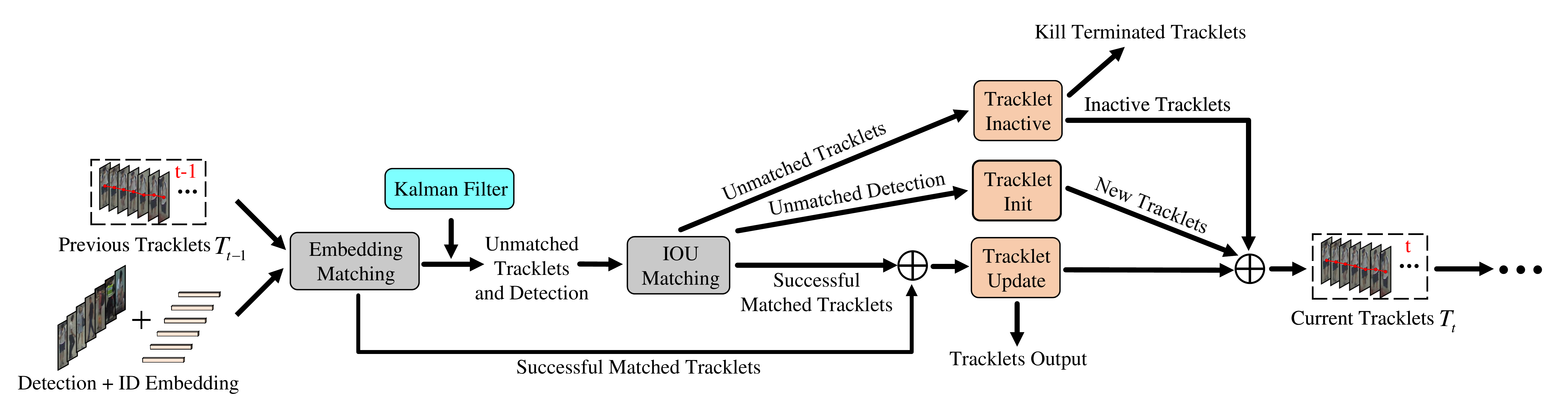}
\end{minipage}
\caption{The details of online tracking. For the detected candidate boxes, we will link them with existing tracklets by a cascade matching design following JDE~\cite{JDE}.}
\label{fig:matching}
\vspace{-0.5em}
\end{figure*}

\subsection{Training details}
\label{sec:training}
For jointly optimizing the object detection and ReID tasks, we minimize a weighted sum of detection losses (consisting of classification loss and box regression loss) and ReID loss. Specifically, the ground-truth annotation for an input image is denoted as $\{B_i\}$, where $B_i=(x^{(i)},y^{(i)},w^{(i)},h^{(i)},c^{(i)})$. Here, $(x^{(i)},y^{(i)})$ indicate the coordinates of the center of bounding box and $(w^{(i)},h^{(i)})$ indicate the size of bounding box. $c^{(i)}$ is the ID index that the object belongs to the $i^{th}$ pedestrian. $C$ is the number of ID in training dataset, which denotes the number of pedestrians. For each location $(x,y)$ on the detection result maps of different resolution, we represent it as a 5D vector $\boldsymbol{t}=(x^{*},y^{*},w^{*},h^{*},p)$, following the standard anchor-based protocol~\cite{yolov3}. Here $(x^{*},y^{*},w^{*},h^{*})$ are the bounding box predictions and $p$ indicates the foreground probability of the bounding box. For foreground and background classification, we define the anchor points in the location  $(x,y)=(\left\lfloor\frac{x^{(i)}}{r}\right\rfloor,\left\lfloor\frac{y^{(i)}}{r}\right\rfloor)$ as positive samples ($r$ is the downsample ratio, \textit{i.e.,} 8, 16, 32, and $\left\lfloor \cdot \right\rfloor$ indicates rounding down to an integer). Therefore, the classification loss~\cite{retinanet} can be formulated as:
\begin{equation}
\mathcal{L}_{cls}(p_t)=-\alpha\left(1-p_{t}\right)^{\gamma} \log \left(p_{t}\right),
\end{equation}
where 
\begin{equation}
p_{t}=\left\{\begin{array}{ll}
p & \text { if } (x,y)=(\left\lfloor\frac{x^{(i)}}{r}\right\rfloor,\left\lfloor\frac{y^{(i)}}{r}\right\rfloor) \\
1-p & \text { otherwise }
\end{array}\right..
\end{equation}
We set $\alpha=0.25$ and $\gamma=0$ following YOLO-v5. For bounding box regression, we employ the CIOU loss~\cite{diou}, which can be defined as, 
\begin{equation}
\mathcal{L}_{reg}(\hat{b}_{x, y})=\left\{\begin{array}{ll}
1-\mathbb{C}(b_i,\hat{b}_{x, y}) & \text{ if } (x,y)=(\left\lfloor\frac{x^{(i)}}{r}\right\rfloor,\left\lfloor\frac{y^{(i)}}{r}\right\rfloor) \\
0 & \text { otherwise }
\end{array}\right..
\end{equation}
where $\mathbb{C}$ represents the CIOU operation. $b_i$ indicates the ground-truth box and $\hat{b}_{x, y}$ is the box prediction at location $(x,y)$ of feature maps. We define our detection loss as follows,
 \begin{equation}
\mathcal{L}_{det}=\frac{1}{N_{p o s}} \sum_{i}^{M} \sum_{x, y}\left(\mathcal{L}_{c l s}\left(p_{x, y}^{i}\right)+\beta \mathcal{L}_{r e g}\left(\hat{b}_{x, y}^{i}\right)\right),
\end{equation}
where $M$ is the number of the resolutions (\textit{i.e.,} 3 in our model) and $N_{p o s}$ denotes the number of positive samples. $\beta$ indicates the loss weight. We set $\beta=0.05$ as that in YOLO-v5.

The definition and training of ReID loss follow the baseline tracker JDE~\cite{JDE}, which uses the cross-entropy loss as the objective for ID embedding learning. In this work, we model this task as a classification task, and use a \textit{FC} layer to map ID embeddings to a class distribution vector $P=\{p(c), c \in [1,2,\dots,C]\}$. By comparing with the one-hot representation of the GT class label $Y^{i}(c) \in \mathbb{R}^{C \times 1 \times 1}$, we compute the ReID loss as
\begin{equation}
L_{id}=-\frac{1}{N} \sum_{i=1}^{N} \sum_{c=1}^{C} Y^{i}(c) \log (p(c)),
\end{equation}
where $N$ denotes the number of objects in the current image.

The joint loss can be written as a weighted linear sum of detection loss $L_{det}$ and ReID loss $L_{id}$, as
\begin{equation}
\mathcal{L}_{total}=\mathcal{L}_{det}+\eta \mathcal{L}_{id},
\end{equation}
where $\eta$ is experimentally set to 0.02 to balance object detection and ReID tasks.

\subsection{Online Tracking}
\label{sec:inference}
In this section, we present how to link the detected candidate boxes with existing tracklets. For a fair comparison, we follow the cascade matching design in JDE~\cite{JDE}, as shown in Fig.~\ref{fig:matching}. Specifically, we first calculate the similarity of ID embeddings between candidate boxes and templates (\textit{i.e.}, the ID embeddings of the previous tracklets) with the cosine metric. Then a Kalman filter is used to provide distance restriction, which removes unreasonable matches between candidate boxes and existing tracklets. After that, the Hungarian algorithm is employed to find the optimal bipartite matching. In the second matching stage, the unmatched tracklets will be re-checked again by an IOU matching, 
as in Sort~\cite{sort}. For the successful matched tracklets, the templates will be updated to adapt to appearance variations, as
\begin{equation}
t_{i}^{t}=\varepsilon t_{i}^{t-1}+(1-\varepsilon) e_{i}^{t},
\end{equation}
where $e_{i}^{t}$ is an ID embedding of candidate box and $i$ indicates the ID index of target. $t_{i}^{t-1}$ and $t_{i}^{t}$ represent the corresponding templates before and after update, respectively. $\varepsilon$ is a weight term, which is set as 0.9, following JDE~\cite{JDE}. The unmatched candidate boxes will be initialized as new tracklets, where the corresponding ID embeddings are simultaneously initialized as the template of new tracklets. The unmatched tracklets are set as the `inactive' state. If a tracklet cannot find its matched box for 30 frames, it will be terminated. On the contrary, the unmatched tracklet will be restored to`active' state when it is successfully matched with a candidate box before terminating. The successful matched tracklets, the new tracklets, and the 'inactive' tracklets will be used to link with the candidate boxes in the next frame.

\section{Experiments}
\label{sec:exp}
To demonstrate the advantages of the proposed CSTrack, we firstly study the effectiveness of each component in our tracking framework in Sec.~\ref{sec:ablations}. Then we compare our method with state-of-the-art approaches in Sec.~\ref{sec:Comparisons}. After that, we further analyze the data association ability of our framework in Sec.~\ref{upper}. Finally, we visualize the qualitative results of CSTrack and analyze the failure cases.

\subsection{Implementation Details}
\label{sec:Imple}

\textbf{Training.} For a fair comparison, we use the same training datasets as the baseline tracker JDE~\cite{JDE}, consisting of ETH~\cite{eth}, CityPerson~\cite{cityperson}, CalTech~\cite{CalTech}, MOT17~\cite{mot16}, CUDK-SYSU~\cite{CUDK} and PRW~\cite{PRW}. Wherein, ETH and CityPerson only provide box annotations, so we only use them to train the detection branch. The other four datasets provide both detection and ID annotations, allowing us to train both object detection and ReID tasks. We chop off the videos in ETH \cite{eth} that are overlapped with the testing benchmark MOT16 \cite{mot16}. For further improving performance, we introduce CrowdHuman \cite{crowdhuman} into training. We train our network with the SGD optimizer for 30 epochs on a single RTX 2080 Ti GPU. The batch size is set to 8. The initial learning rate is $5 \times 10^{-4}$, and we decay the learning rate to $5 \times 10^{-5}$ at the 20th epoch. The hyperparameters of the object detector are set the same as Yolo-v5 and others follow IDE~\cite{JDE}. The \textit{Backbone} and \textit{Neck} networks are initialized with the parameters pre-trained on COCO~\cite{coco}. For the reciprocal network, we set the feature size after \textit{avg-pooling} layer as $(H^{\prime},W^{\prime}) = (6,10)$ and initialize the trainable parameters $\lambda_{1/2}=0.5$.

\textbf{Testing and Evaluation Metrics.} We evaluate our network on MOT16~\cite{mot16}, MOT17~\cite{mot16} and the recently-released MOT20~\cite{mot20}. Specifically, MOT16~\cite{mot16} and MOT17~\cite{mot16} contain the same 7 testing videos, yet some videos are re-annotated. MOT20~\cite{mot20} consists of 4 testing videos under extremely crowded scenes. Following the common practices in MOT Challenge~\footnote{\url{https://motchallenge.net}}, we adopt the CLEAR metric~\cite{evaluating}, particularly Multiple Object Tracking Accuracy (MOTA) to evaluate the overall performance. ID F1 Score (IDF1)~\cite{idf1} is  employed to evaluate the association performance. We also report other metrics including the Most Tracked ratio (MT) for the ratio of most tracked ($>80 \%$ time) objects, Most Lost ratio (ML) for most lost ($<20 \%$ time) objects, the number of Identity Switch (ID Sw.)~\cite{idsw} and FPS for measuring frame rate of the overall system. 

Our tracker is implemented using Python 3.7 and PyTorch 1.6.0. The experiments are conducted on a single RTX 2080Ti GPU and Xeon Gold 5218 2.30GHz CPU. 

\begin{table}[!t]
 	\vspace{-1em}  
    \begin{center}

        \caption{Component-wise Analysis of CSTrack on MOT16 testing set.} 
        \vspace{0.5em}
        \fontsize{8pt}{4.3mm}\selectfont
        \begin{threeparttable}
            \begin{tabular}{@{}c@{} | @{}c@{} | @{}c@{}  @{}c@{} | @{}c@{}  @{}c@{} @{}c@{}}
                \cline{1-7}
                NUM~~&~~~~~Method~~~~~&~~REN~~&~~SAAN~~& ~MOTA$\uparrow$~  & ~IDF1$\uparrow$~  & ~ID Sw.$\downarrow$~
                \\
                \cline{1-7} 
\ding{172} & JDE(yolo-v3)&            &            & 64.4 & 55.8 & 1544\\
\ding{173} & JDE(yolo-v5)&            &            & 68.9 & 60.7 & 1798 \\
\ding{174} & JDE(yolo-v5)& \checkmark &            & 70.8 & 63.1 & 1365\\
\ding{175} & JDE(yolo-v5)&            & \checkmark & 69.4 & 69.3 & 1226\\
\ding{176} & JDE(yolo-v5)& \checkmark & \checkmark & \bf{72.9} & \bf{71.6} & \bf{1121} \\
                \cline{1-7}

            \end{tabular}

        \end{threeparttable}
        \vspace{-2em}
        \label{tab:ablation}
    \end{center}
\end{table}

\subsection{Ablation Studies~\protect\footnote{Due to the limited submissions of the MOT Challenge (totally 4 times per model on each benchmark), some ablation studies are verified on validation set.}}
\label{sec:ablations}

\textbf{Component-wise Analysis.} In this section, we verify the effectiveness of our proposed reciprocal network (REN in Sec.~\ref{sec:REN}) and scale-aware attention network (SAAN in Sec.~\ref{sec:SAAN}). For a fair comparison with JDE, all experiments are trained using the six datasets mentioned above and validated in the MOT16 testing set~\cite{mot16}. The comparison experimental results are presented in Tab.~\ref{tab:ablation}. We first replace the object detector in JDE~\cite{JDE} from YOLO-v3~\cite{yolov3} to YOLO-v5, which ensures a better performance and faster inference speed (\ding{172} \textit{vs.} \ding{173}). The YOLO-v5 replacement provides a strong baseline for our following design. When equipped with the proposed reciprocal network (REN), it achieves 1.9 points gains on MOTA, 2.4 points gains on IDF1 and the IDs decrease from 1798 to 1365 (\ding{173} \textit{vs.} \ding{174}). This demonstrates the effectiveness of our REN component in learning task-dependent representations to improve both detection and association performance. The introduced SAAN module aims to avoid the semantic gap when matching ID embeddings, and it brings 8.6 points gains on IDF1 (\ding{174} \textit{vs.} \ding{176}). Compared with the vanilla JDE~\cite{JDE}, our tracker significantly improves tracking performance (\ding{172} \textit{vs.} \ding{176}), \textit{i.e.,} MOTA +8.5\%, IDF1 +15.8\% and ID Sw. decreased from 1544 to 1121.

\begin{table}[!t]
    \begin{center}

        \caption{Ablation study of REN on MOT17 validation set. SR represents self-relation, CR represents cross-relation.} 
        \vspace{0.5em}
        \fontsize{8pt}{4.5mm}\selectfont
        \begin{threeparttable}
            \begin{tabular}{@{}c@{}| @{}c@{}  @{}c@{} | @{}c@{}  @{}c@{} | @{}c@{}  @{}c@{} @{}c@{} @{}c@{}}
                \cline{1-8}
                NUM~~&~~SR~~&~~CR~~&~~~~~$\lambda_1$~~~~~&~~~~~$\lambda_2$~~~~~& ~MOTA$\uparrow$~  & ~IDF1$\uparrow$~  & ~ID Sw.$\downarrow$~\\
                \cline{1-8} 
\ding{172} & \checkmark &            & $-$ & $-$ &  65.3 & 69.5  & 713  \\
\ding{173} &            & \checkmark & $-$ & $-$ & 64.8 & 67.2 & 616        \\
\ding{174} & \checkmark & \checkmark & 0.5 & 0.5 & 64.5  & 68.9  & 679  \\
\ding{175} & \checkmark & \checkmark & 0.12122 & 0.31519 & \bf{66.0} & \bf{70.7} & \bf{535} \\
                \cline{1-8}
\ding{176} &  \multicolumn{4}{c|}{w/o REN} & 64.6  & 68.7  &  739  \\
                \cline{1-8}
\ding{177} &  \multicolumn{4}{c|}{replace REN with CNN} & 64.9  & 69.0  &  717 \\
                \cline{1-8}

            \end{tabular}

        \end{threeparttable}
        \vspace{-1em}
        \label{tab:REN}
    \end{center}
\end{table}

\begin{table}[!t]
    \begin{center}

        \caption{Impact of attention module arrangement in our tracker, where the module will be arranged serially, as~\cite{cbam}. } 
        \vspace{0.5em}
        \fontsize{8pt}{4.3mm}\selectfont
        \begin{threeparttable}
            \begin{tabular}{@{}c@{} | @{}c@{}  @{}c@{} | @{}c@{}  @{}c@{} | @{}c@{} @{}c@{} @{}c@{}}
                \cline{1-8}
\multirowcell{2}{NUM}~~& \multicolumn{2}{c|}{Each resolution} & \multicolumn{2}{c|}{concatenate}  &~~\multirowcell{2}{MOTA$\uparrow$}~~&~~\multirowcell{2}{IDF1$\uparrow$} ~~&~~\multirowcell{2}{ID Sw.$\downarrow$}~~\\ 
 &~~SAM~&~CAM~&~~SAM~~&~~CAM~~& &  & \\ \hline
    
                \cline{1-8} 
\ding{172} &  &  &  &  & 64.1 & 66.4 & 723 \\
\ding{173} & \checkmark &  &  &  & 66.0 & 70.4 & 633  \\
\ding{174} &  & \checkmark &  &  & 65.4 & 68.4 & 659 \\
\ding{175} & \checkmark & \checkmark &  &  & 64.9 & 69.8 & 690 \\
\ding{176} & \checkmark &  & \checkmark &  & 64.1 & 69.3 & 721 \\
\ding{177} & \checkmark &  &  & \checkmark & \bf{66.0} & \bf{70.7} & \bf{535} \\
\ding{178} & \checkmark &  & \checkmark & \checkmark & 64.3 & 69.9 & 700 \\
\cline{1-8}
\ding{179} & \multicolumn{4}{c|}{Basic ReID Head (w/o SAAN)} & 63.8 & 61.3 & 1047 \\
                \cline{1-8}

            \end{tabular}
        \end{threeparttable}
        \vspace{-1em}
        \label{tab:attention}
    \end{center}
\end{table}

\begin{table}[!t]
    \begin{center}

        \caption{Analysis of training dataset settings on MOT16 testing set. ``S'' indicates only using MOT17 for training. ``M'' indicates using six datasets mentioned in the manuscript. ``B'' denotes training on the seven datasets including ``M'' and Crowdhuman dataset. It verifies that CSTrack can achieve efficient performance, even only using MOT17 for training.} 
        \vspace{0.5em}
        \fontsize{8pt}{4.3mm}\selectfont
        \begin{threeparttable}
            \begin{tabular}{@{}c@{} | @{}c@{} | @{}c@{}  @{}c@{}  @{}c@{} | @{}c@{} @{}c@{} @{}c@{}}
                \cline{1-8}
NUM~~&~~~~~~Settings~~~~~~&~~image~~&~~bbox~~&~~ID~~~&~MOTA$\uparrow$~&~IDF1$\uparrow$~&~ID Sw.$\downarrow$
                \\
                \cline{1-8} 
\ding{172} & JDE (M) & 54K & 270K & 8.7K & 64.4 & 55.8 & 1544  \\
\ding{173} & CSTrack (S) & 5K & 112K & 0.5K & 71.3 & 68.6 & 1356  \\
\ding{174} & CSTrack (M) & 54K & 270K & 8.7K & 72.9 & 71.6 & \bf{1050} \\
\ding{175} & CSTrack (B) & 73K & 740K & 8.7K& \bf{75.6} & \bf{73.3} & 1121 \\
                \cline{1-8}

            \end{tabular}
        \end{threeparttable}
        \vspace{-1em}
        \label{tab:trainset}
    \end{center}
\end{table}

\begin{table}[!t]
    \begin{center}

        \caption{Impact of ID loss weight setting in our tracker. $\eta$ indicates the ID loss weight. }
        \vspace{0.5em}
        \fontsize{8.5pt}{4.3mm}\selectfont
        \begin{threeparttable}
            \begin{tabular}{@{}c@{} | @{}c@{} | @{}c@{} @{}c@{} @{}c@{}}
                \cline{1-5}
                NUM~~&~~~~~~$\eta$~~~~~~& ~~~MOTA$\uparrow$~~~  & ~~~IDF1$\uparrow$~~~&~~~ID Sw.$\downarrow$~~~ \\
                \cline{1-5} 
\ding{172} & 1 & 21.1 & 35.2 & 966 \\
\ding{173} & 0.8 & 33.5 & 45.7 & 1002  \\
\ding{174} & 0.6 & 54.2 & 60.5 & 693 \\
\ding{175} & 0.4 & 55.0 & 62.3 & 683 \\
\ding{176} & 0.2 & 64.5 & 69.0 & 713 \\
\ding{177} & 0.1 & 65.3 & 69.1 & 642 \\
\ding{178} & 0.05 & 65.2 & 70.3 & 652 \\
\ding{179} & 0.02 & 66.0 & \bf{70.7} & \bf{535} \\
\ding{180} & 0.01 & \bf{66.2} & 68.6 & 701 \\
\ding{181} & 0.005 & 65.8 & 67.2 & 742 \\
                \cline{1-5}
            \end{tabular}
        \end{threeparttable}
        \vspace{-2em}
        \label{tab:Idloss}
    \end{center}
\end{table}

\begin{table*}[!t]
    \begin{center}

        \caption{Comparison with state-of-the-art online MOT trackers on the MOT16, MOT17 and MOT20 benchmarks. $\#$ indicates one-shot method. $*$ indicates other joint detection and tracking methods which adopt non-ReID methods for data association. Others without special sign indicate two-stage methods. FP and FN mean false positive and false negative, respectively. FPS denotes frame rates for the whole tracking framework. ``CSTrack-S'' indicates our lightweight version (detailed in Sec.~\ref{sec:exp}).} 
        \vspace{0.5em}
        \fontsize{9pt}{4.3mm}\selectfont
        \begin{threeparttable}
            \begin{tabular}{@{}c@{} | @{}c@{} | @{}c@{} | @{}c@{}  @{}c@{}  @{}c@{} @{}c@{} @{}c@{} @{}c@{} @{}c@{} @{}c@{} @{}c@{}}
                \cline{1-11}
~~~~~Dataset~~~~~~&~~~~~~~~~~~~~Method~~~~~~~~~~~~~&~~~~~Published~~~~~&~~~MOTA$\uparrow$~~~&~~~IDF1$\uparrow$~~~&~~~MT$\uparrow$~~~&~~~ML$\downarrow$~~~&~~~FP$\downarrow$~~~&~~FN$\downarrow$~~&~~ID Sw.$\downarrow$~~&~~FPS$\uparrow$
                \\
                \cline{1-11} 
\multirowcell{16}{MOT16} & SORT$\_$POI~\cite{sort} & ICIP2016  & 59.8 & 53.8 & 25.4 & 22.7 & 8698 & 63245 & 1423  & $\textless$8.6 \\
      & POI~\cite{poi}     & ECCV2016  & 66.1 & 65.1 & 34.0 & 21.3 & 5061 & 55914 & 805  & $\textless$5.2 \\
      & DeepSORT-2~\cite{deepsort} & ICIP2017  & 61.4 & 62.2 & 32.8 & 18.2 & 12852 & 56668 & 781  & $\textless$6.7 \\
      & RAN~\cite{RAN}   & WACV2018  & 63.0 & 63.8 & 39.9 & 22.1 & 13663 & 53248 &{\bf482}  & $\textless$1.5 \\
      & TAP~\cite{HOGM}   &ICPR2018   & 64.8 & \bf{73.5} & 38.5 & 21.6 & 12980 & 50635 & 794  & $\textless$8.0 \\
      & IAT\cite{IAT}   & WACV2019   & 48.8  & - & 15.8 & 38.1 & 5875 & 86567 & 936  & - \\
      & CNNMTT\cite{CNNMTT}   & MTA2019   & 65.2 & 62.2 & 32.4 & 21.3 & 6578 & 55896 & 946  & $\textless$5.2 \\
     &$*$Tracktor++~\cite{tracktor} &ICCV2019 & 56.2 & 54.9 & 20.7 & 35.8 & \bf{2394} & 76844 & 617 & 1.8\\
     & STPP\cite{STPP}   & TNNLS2020   & 50.5  & - & 19.6 & 39.4 & 5939 & 83694 & 638  & - \\
     & TPM\cite{TPM}   & PR2020   & 50.9  & - & 19.4 & 39.4 & 4866 & 84022 & 619  & - \\
     & $*$TubeTK$\_$POI~\cite{tubetk}  & CVPR2020  & 66.9 & 62.2 & 39.0 & 16.1 & 11544 & 47502 & 1236 & 1.0  \\
     & $*$CTrackerV1~\cite{CTrack} & ECCV2020 & 67.6 & 57.2 & 32.9 & 23.1 & 8934 & 48305 & 1897 & 6.8  \\
      & $\#$JDE~\cite{JDE}   & ECCV2020  & 64.4 & 55.8 & 35.4 & 20.0 &  10642 & 52523 & 1544 & 18.8 \\
      & $\#$FairMOTv2~\cite{fairmot} & IJCV2021 & 74.9 & 72.8 & \bf{44.7} & \bf{15.9} & 10163 & 34484 & 1074  & 18.9  \\
      & $\#$CSTrack & Ours & \bf{75.6} & 73.3 & 42.8 & 16.5 & 9646 & \bf{33777} & 1121 & 16.4 \\
      & $\#$CSTrack-S & Ours & 73.8 & 71.0 & 40.1 & 18.7 & 9157 & 37329 & 1313 &  \bf{34.6}  \\
\midrule  
\multirowcell{9}{MOT17} &$*$Tracktor++~\cite{tracktor} &ICCV2019 & 56.3 & 55.1 & 21.1 & 35.3 & \bf{8866} & 235449 & \bf{1987} & 1.8\\
      & STPP\cite{STPP}   & TNNLS2020   & 52.4  & - & 22.4 & 40.0 & 20176 & 246158 & 2224  & - \\
      & TPM\cite{TPM}     & PR2020   & 52.4  & - & 22.4 & 40.0 & 19922 & 246183 & 2215  & - \\
      & $*$TubeTK~\cite{tubetk}  & CVPR2020  & 63.0 & 58.6 & 31.2 & 19.9 & 27060 & 177483 & 4137 & 3.0 \\
      & $*$CTrackerV1~\cite{CTrack} & ECCV2020 & 66.6 & 57.4 & 32.2 & 24.2 & 22284 & 160491 & 5529 & 6.8 \\
      & $*$CenterTrack~\cite{centertrack} & ECCV2020 & 67.8 & 64.7 & 34.6 & 24.6 & 18498 & 160332 & 3039 & 22.0 \\
      & $\#$JDE~\cite{JDE}   & ECCV2020  & 63.0 & 59.5 & 35.7 & 17.3 &  39888 & 162927 & 6171 & 18.8 \\
      & $\#$FairMOTv2~\cite{fairmot} & IJCV2021 & 73.7 & \bf{72.3} & \bf{43.2} & \bf{17.3} & 27507 & 117477 & 3303 & 18.9 \\
      & $\#$CSTrack  & Ours  & \bf{74.9} & \bf{72.3} & 41.5 & 17.5 & 23847 & \bf{114303} & 3567 & 16.4 \\
      & $\#$CSTrack-S & Ours & 72.9 & 70.3 & 38.7 & 18.9 & 23148 & 125418 & 4119 & \bf{34.6} \\
\midrule  
\multirowcell{4}{MOT20} & $\#$JDE~\cite{JDE}  & ECCV2020  & 40.4 & 33.5 & 8.9 & 25.3 &  24228 & 271853 & 12301 & 6.7\\
      & $\#$FairMOTv2~\cite{fairmot} & IJCV2021 & 61.8 & 67.3 & \bf{68.8} & \bf{7.6} & 103440 & \bf{88901} & 5243 & 8.9 \\
      & $\#$CSTrack  & Ours & \bf{66.6} & \bf{68.6} & 50.4 & 15.5 & 25404 & 144358 & \bf{3196} & 4.5 \\
      & $\#$CSTrack-S & Ours & 64.2 & 65.0 & 47.4 & 18.4 & \bf{21088} & 160253 & 3854 & \bf{12.4} \\
                \cline{1-11}

            \end{tabular}
        \end{threeparttable}
        \vspace{-1.5em}
        \label{tab:comparison}
    \end{center}
\end{table*}

\textbf{Self-relation and Cross-relation Analysis.} To verify the effectiveness of self-relation and cross-relation in our tracking framework, we conduct experiments with different settings and present the comparison results in Tab.~\ref{tab:REN}. We train all the models on half of the training set and validate on another half. Compared with the model without REN (\ding{172} \textit{vs.} \ding{176}), employing self-relation to learn specificities can bring 0.7\% gains on MOTA and 0.8\% gains on IDF1, respectively. When only introducing the cross-relation, our method sightly outperforms the model without REN by 0.2\% on MOTA, but IDF1 decreases from 68.7\% to 67.2\% (\ding{173} \textit{vs.} \ding{176}). The reason for this degradation is that only learning commonalities is insufficient to resolve the underlying competition between object detection and ReID. When we merge self-relation and cross-relation with equal weights (\textit{i.e.,} we set $\lambda_{1/2}$ as constant weights 0.5, see \ding{174}), the performance is inferior to the model only introducing self-relation (\ding{172} \textit{vs.} \ding{174}). But surprisingly, when introducing learning weights to adjust the fusion of self-relation and cross-relation, we obtain promising results with MOTA of 66.0\%  and IDF1 of 70.7\%. Notably, the ID Sw. decreased about 25\% (\ding{172} \textit{vs.} \ding{175}), demonstrating the effectiveness of our reciprocal network. It is noted that the learning weights $\lambda_1$ and $\lambda_2$ converge to 0.12122 and 0.31519, respectively. This reveals that different tasks have different requirements for features. Meanwhile, it verifies the specificities and commonalities learning can efficiently learn task-dependent representations, which improves the performance remarkably. What's more, to further verify the effectiveness of our reciprocal network, we replace REN with CNN module which has a nearly similar computational cost (74.6M and 1304 GFlops for REN version \textit{v.s.} 77.2M and 1308 GFLOPs for CNN version). We observe that REN can achieve better results, even with fewer parameters and FLOPs (\ding{175} \textit{vs.} \ding{177}).


\textbf{Impact of Attention Module Arrangement.} In order to understand the impact of the attention module arrangement in SAAN (see Sec.~\ref{sec:SAAN}), we conduct ablation studies on MOT17 validation set~\cite{mot16}.  As shown in Tab.~\ref{tab:attention}, compared with one-shot tracker equipped with the basic ReID head, our model achieves better MOTA and IDF1 (\ding{172}$\sim$\ding{178} \textit{vs.} \ding{179}). This verifies the effectiveness of our designed SAAN network in learning discriminative embeddings. As our core motivation is to robustly aggregate information from different resolutions, we validate our model with different attention module arrangements. Firstly, we configure different combinations of spatial attention modules (SAM) and channel attention modules (CAM) at each resolution. The results show that introducing spatial attention outperforms the basic model by +1.9\% in MOTA, +1.0\% in IDF1 and ID Sw. decreased from 723 to 633, respectively (see \ding{172}$\sim$\ding{175}). This demonstrates that SAM module is useful to enhance the target-related features and suppress background noise. Considering results from \ding{176} to \ding{178}, we observe that jointly applying channel attention in aggregated feature can achieve even better results, further decreasing ID Sw. from 633 to 535.

\textbf{Training Dataset Setting.} We conduct another ablation study to evaluate the impact of different training dataset setting, and the corresponding results are presented in Tab.~\ref{tab:trainset}. According to the results of Tab.~\ref{tab:trainset}, when only using the MOT17 dataset~\cite{mot16} for training (see \ding{173}), our tracker achieves 71.3 MOTA and 68.6 IDF1, which outperforms many existing trackers. For a fair comparison, we employ the same setting as our baseline tracker JDE~\cite{JDE}, \textit{i.e.,} training on the six datasets mentioned in Sec.~\ref{sec:Imple}.
From the result presented in Tab.~\ref{tab:trainset} (\ding{172} \textit{vs.} \ding{174}), our model surpasses JDE by 8.5 point in MOTA and 15.8 point in IDF1, and the ID Sw. decreased form 1544 to 1050. It further verifies that our design is simple and effective to improve tracking performance by alleviating the competition between object detection and ReID in the one-shot tracker. Moreover, we further add CrowdHuman~\cite{crowdhuman} images into training, and observe that it brings gains of 2.7 point on MOTA and 1.7 points on IDF1 (\ding{174} \textit{vs.} \ding{175}). One possible reason is that training with CrowdHuman dataset can improve model tracking performance in crowded scene.

\textbf{Impact of ID Loss Weight.} To jointly optimize object detection and target representation (\textit{i.e.} ID embedding) learning, we balance their loss with a handcrafted weight $\eta$. To study the impact of $\eta$ for ID embedding learning, an ablation experiment is conducted and the results are shown in Tab.~\ref{tab:Idloss}. \ding{172}$\sim$\ding{176} demonstrate that a large ID loss weight $\eta$ will degrade detection performance (see MOTA score), while a small ID loss weight degrades matching performance (see IDF1 and ID Sw. score). We set $\eta$ to 0.02 considering better scoring across different evaluation criteria.

\subsection{Comparisons with State-of-the-art Trackers} 
\label{sec:Comparisons}
We compare our multi-object tracker CSTrack with other start-of-the-art MOT tracking methods on MOT16~\cite{mot16}, MOT17~\cite{mot16} and MOT20~\cite{mot20} benchmarks. The compared methods can be divided into three categories. The first one is two-stage method including SORT$\_$POI~\cite{sort}, POI~\cite{poi}, DeepSORT-2~\cite{deepsort}, RAN~\cite{RAN}, TAP~\cite{HOGM}, CNNMTT\cite{CNNMTT}, IAT~\cite{IAT}, STPP~\cite{STPP} and TPM~\cite{TPM}. The second one is the one-shot method including JDE~\cite{JDE} and FairMOTv2~\cite{fairmot}. The third one is the other joint detection and tracking method without using ReID for data association, including CTrackerV1~\cite{CTrack}, TubeTK~\cite{tubetk}, Tracktor++~\cite{tracktor} and CenterTrack~\cite{centertrack}. The evaluations of all above methods on these benchmarks are performed by the official online server~\footnote{\url{https://motchallenge.net}}. 

\textbf{Two-stage Methods.}  The comparison with two-stage methods follows the protocols in~\cite{mot16}. All the two-stage methods use the private detector (\textit{i.e.,} FasterRCNN~\cite{fasterrcnn}), which is trained on a large private pedestrian detection dataset. The main differences among them lie in their ID embedding extraction and association strategies. As shown in Tab.~\ref{tab:comparison}, our proposed CSTrack outperforms the prior state-of-the-art methods by a significant margin and performs much faster. For example, comparing with the top performance two-stage tracker (\textit{i.e.,} POI~\cite{poi}), CSTrack outperforms it by +9.5\% on MOTA and +8.2\% on IDF1. Considering the inference speed, the proposed CSTrack runs faster than existing two-stage methods, \textit{i.e.,} 16.4 FPS \textit{vs}. 0.5$\sim$8.6 FPS on MOT16~\cite{mot16}. Furthermore, we construct a light version, namely CSTrack-S, which reduces convolution channels and layers of CSTrack to 30\% of its original settings. For the lightweight vision, the running speed is clearly sped up to 34.6 FPS with only a small performance drop (MOTA -1.8$\sim$2.4\% and IDF1 -2.3$\sim$3.6\%). In the term of data association, it is not feasible to directly compare these methods because of the nonuniform object detectors. Therefore, we further analyze the upper bound of association under the same detections input in Sec.~\ref{upper}.

\textbf{One-shot Trackers.}  We present the evaluation results with the comparisons to recently-prevailing one-shot trackers on MOT16~\cite{mot16}, MOT17~\cite{mot16} and MOT20~\cite{mot20}. Our CSTrack achieves MOTA score of 75.6 and IDF1 score of 73.3 on MOT16, which evidently outperforms JDE~\cite{JDE} on MOTA of +11.2\% and IDF1 of +17.5\%. Besides, we compare our model with the recently-proposed one-shot FairMOTv2~\cite{fairmot}, which achieves the second performance of MOTA in all three benchmarks. FairMOTv2~\cite{fairmot} is trained on the same datasets as CSTrack, which makes the comparison fair. On MOT16 and MOT17, the false negative (FN) and false positive (FP) of our tracker surpass FairMOTv2 for 5.1\%$\sim$13.3\% on FP and 2.1\%$\sim$2.7\% on FN. On MOT20, our tracker outperforms FairMOTv2 by significantly decreasing FP from 103440 to 25404 and ID Sw. from 5243 to 3196. It means that our tracker can achieve a comparable or even superior IDF1 score when we use fewer boxes, which actually verifies that our method authentically improves the performance of data association.


\textbf{Other Joint Detection and Tracking Methods.} To further evaluate the proposed method, we conduct some comparisons with other joint detection and tracking methods, which adopt non-ReID methods for data association. The evaluation results on MOT16 and MOT17 are presented in Tab.~\ref{tab:comparison}. Compared with these methods, our tracker significantly improves tracking performances, especially on data association ability, \textit{i.e.,} IDF1 +11.1\%$\sim$18.4\% in MOT16 and +7.6\%$\sim$17.2\% in MOT17. It verifies that our design effectively use ID information, which makes the data association ability of our tracker outperform the existing joint detection and tracking methods.

\begin{figure*}[t]
\begin{minipage}[b]{1\linewidth}
\includegraphics[width=0.99\linewidth]{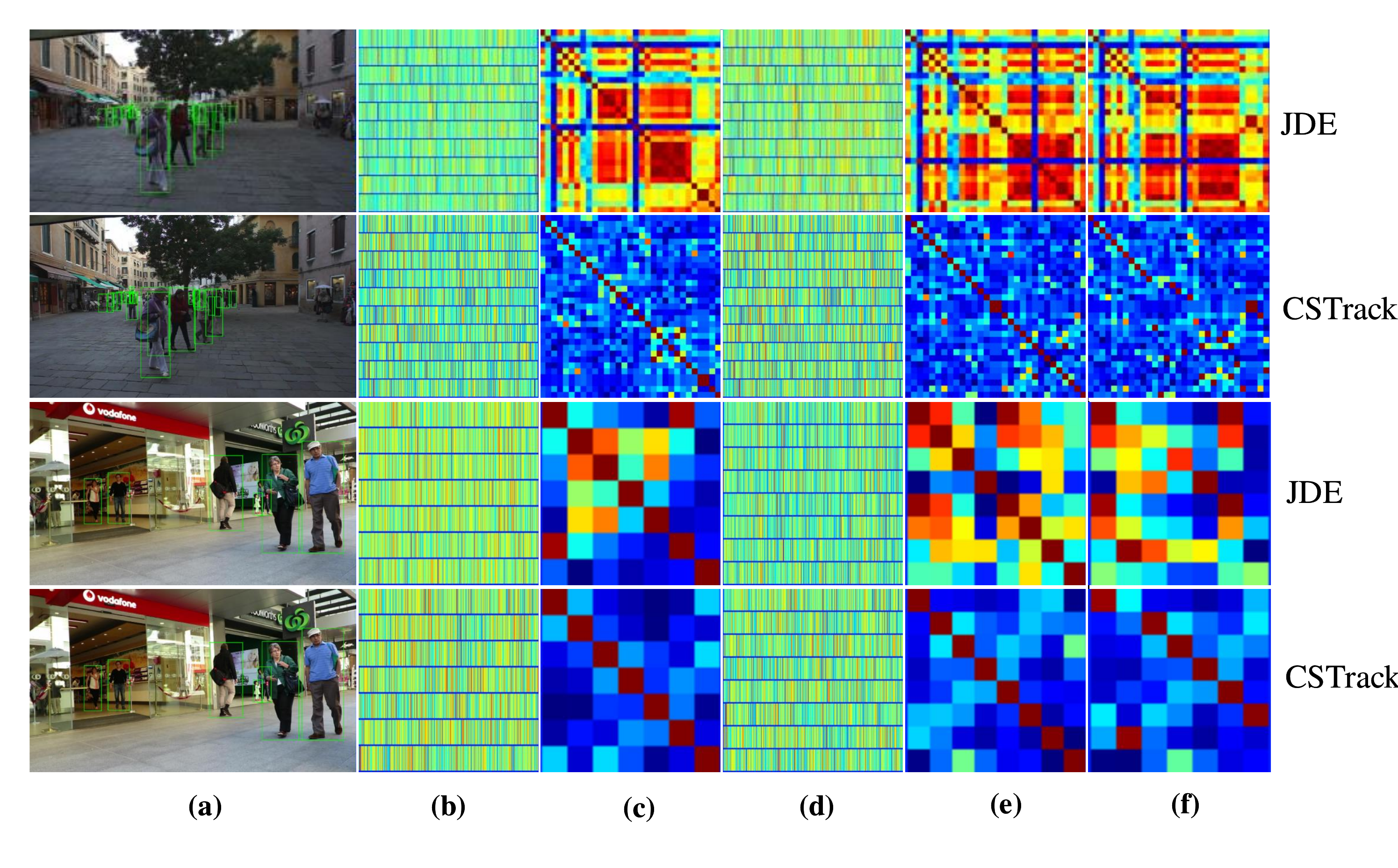}
\vspace{-0.5cm}
\end{minipage}
\caption{Visualization about the discriminative ability of ID embeddings: (a) Example Image with ground truth detection results. (b) ID embeddings of the current example frame. (c) Cosine metric matrix between two ID embeddings of the current example frame. (d) Templates (\textit{i.e.,} ID embeddings of the previous tracklets). (e) Cosine metric matrix between two ID embeddings of the template. (f) Cosine metric matrix between ID embeddings of the current example frame and the template. Red color indicates a higher correlation between two features. Best viewed in color and zoom in.}
\label{fig:reid}
\vspace{-1em}
\end{figure*}

\subsection{Further Analysis}
\label{upper}
\begin{table}[!t]
    \begin{center}

        \caption{Comparison with state-of-the-art trackers on the HiEve banchmarks (private detection).} 
        \vspace{0.5em}
        \fontsize{8pt}{3.5mm}\selectfont
        \begin{threeparttable}
            \begin{tabular}{@{}c@{} @{}c@{}  @{}c@{}  @{}c@{}  @{}c@{} @{}c@{}  @{}c@{} @{}c@{}}
                \cline{1-8}
~~~~~~~Model~~~~~~~&~MOTA$\uparrow$~&~~IDF1$\uparrow$~~&~~MT$\uparrow$~~&~~ML$\downarrow$~~&~~FP$\downarrow$~~&~~FN$\downarrow$~~&~~ID Sw.$\downarrow$
                \\
                \cline{1-8} 
DeepSORT~\cite{deepsort} & 27.1 & 28.5 & 8.4 & 41.4 & 5894 & 42668 & 3122 \\
JDE~\cite{JDE} & 33.1 & 36.0 & 15.1 & 24.1 & 9526 & 33327 & 3605 \\
FairMOT~\cite{fairmot}  & 35.0 & 46.6 & 16.2 & 44.1 & 6523 & 37750 & 2312\\
CenterTrack~\cite{centertrack}  & 42.4 & 38.3 & \bf{23.9} & 26.5 & 5802 & \bf{29766} & 2940\\
NewTracker~\cite{newtracker}  & 46.4 & 43.2 & \bf{26.3} & 30.8 & 4667 & 30489 & 2133\\
CSTrack  & \bf{48.6} & \bf{51.4} & 20.4 & 33.5 & \bf{2366} & 31933 & \bf{1475}\\
                \cline{1-8} 

            \end{tabular}
        \end{threeparttable}
        \label{tab:hieve}
        \vspace{-1.5em} 
    \end{center}
\end{table}

\textbf{HiEve Challenge.}
For further evaluating the strength of CSTrack, we show a comparison on a  more challenging benchmark, \textit{i.e.,} Human in Events dataset (HiEve)~\cite{HiEve}.
HiEve focuses on human-centric complex events, \textit{e.g.,} fighting, quarreling, accident, robbery and shopping, including 19 videos for training and 13 videos for testing. For evaluating on the HiEve benchmark, we fine-tune our method on the training set of HiEve following the same training procedure on MOT challenge benchmarks (See Sec.~\ref{sec:Imple}). All evaluations are performed by the official online server~\footnote{\url{http://humaninevents.org/}}, as shown in Tab.~\ref{tab:hieve}. The proposed CSTrack outperforms its baseline JDE~\cite{JDE} on MOTA for +16.5\% and IDF1 for +15.4\%. Compared with other top-ranked methods, our tracker still shows better tracking performance on MOTA and IDF1.

\begin{table}[!t]
    \begin{center}

        \caption{To show the potential of CSTrack. IDP and IDR describe the precision and recall of matching target tracklets.} 
        \vspace{0.5em}
        \fontsize{9pt}{4.3mm}\selectfont
        \begin{threeparttable}
            \begin{tabular}{@{}c@{} @{}c@{}  @{}c@{}  @{}c@{}  @{}c@{} @{}c@{}}
                \cline{1-6}
~~~~~~~~Model~~~~~~~~&~~MOTA$\uparrow$~~&~~IDF1$\uparrow$~~&~~IDP$\uparrow$~~&~~IDR$\uparrow$~~&~~ID Sw.$\downarrow$
                \\
                \cline{1-6} 
JDE~\cite{JDE} & 97.6 & 87.6 & 88.3 & 86.9 & 871 \\
DeepSORT-2~\cite{deepsort} & \bf{98.9} & 95.6 & 95.9 & 95.3 & \bf{93} \\
CSTrack  & \bf{98.9} & \bf{96.6} & \bf{97.1} & \bf{96.1} & 162 \\
                \cline{1-6}

            \end{tabular}
        \end{threeparttable}
        \vspace{-1.5em}
        \label{tab:upper}
    \end{center}
\end{table}

\textbf{Upper-bound Analysis.} 
In this section, we study the upper-bound of the data association ability of our model. As a common wisdom, tracking performance is greatly influenced by the adopted object detectors~\cite{evaluating}. To eliminate the influence of object detectors, we validate on the MOT16 training set by replacing the detection results with ground-truth bounding boxes. For a fair comparison, all of these methods are trained on the same training set, \textit{i.e.,} MOT17~\cite{mot16}. As shown in Tab.~\ref{tab:upper}, compared with JDE~\cite{JDE}, our method yields substantial improvements on data association reflected by 9 points gains on IDF1 score. Moreover, the IDF1 score of our method surpasses the widely-used two-stage method DeepSORT-2~\cite{deepsort}, which employs Wide Residual Network (WRN)~\cite{wan} as ID embedding extraction model. It further verifies our design can effectively improve association ability of one-shot tracker. 

\begin{figure*}
\centerline{\includegraphics[height=50\baselineskip]{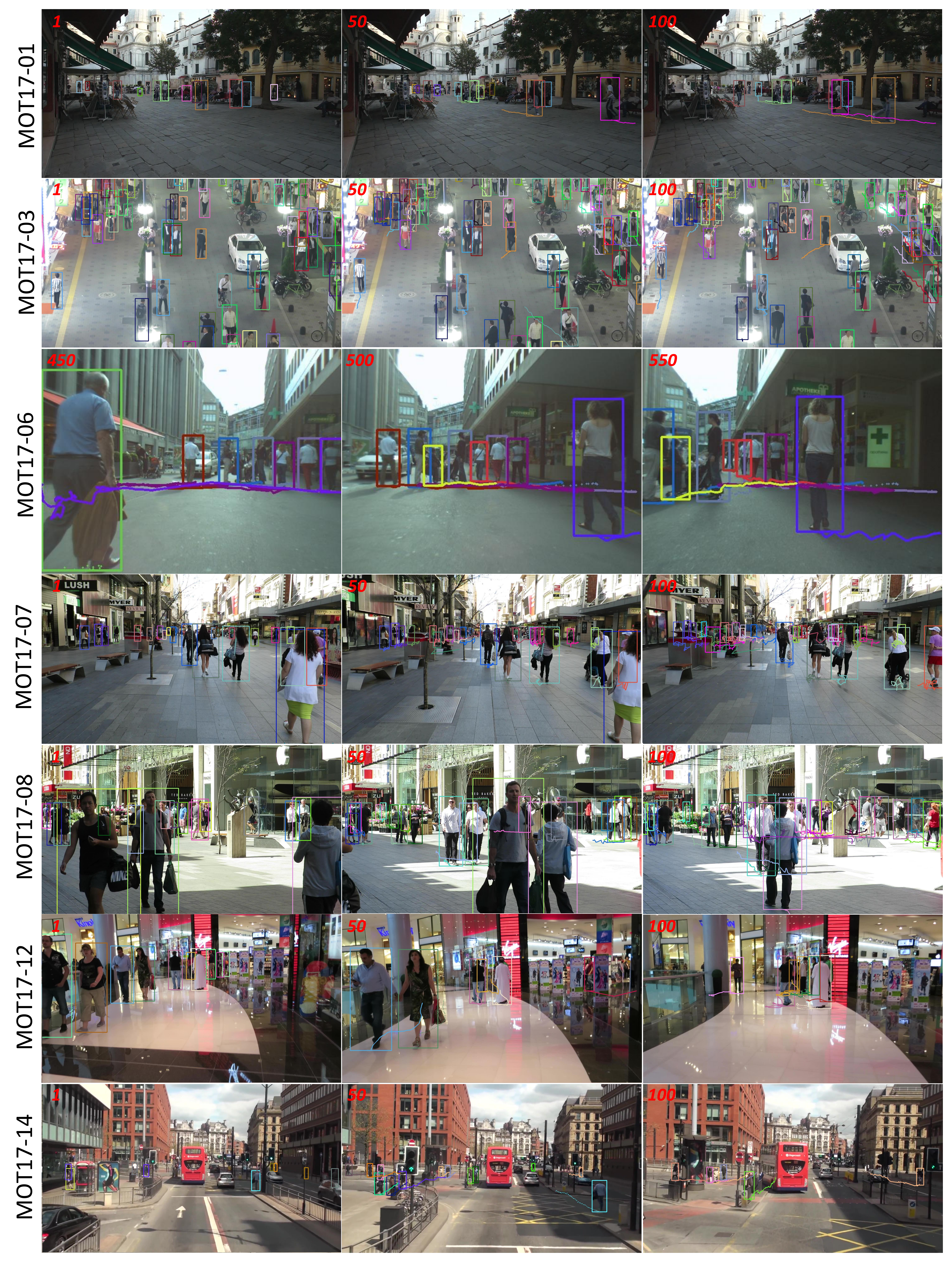}}
\caption{Qualitative results of our CSTrack on MOT17~\cite{mot16}. Different colored bounding boxes denote different identity. The line under each bounding box indicates the tracklet of each target. The frame number is in the upper left of the figure. Best viewed in color and zoom in.}
\label{fig:vis_17}
\end{figure*}

\begin{figure*}
\centerline{\includegraphics[height=27\baselineskip]{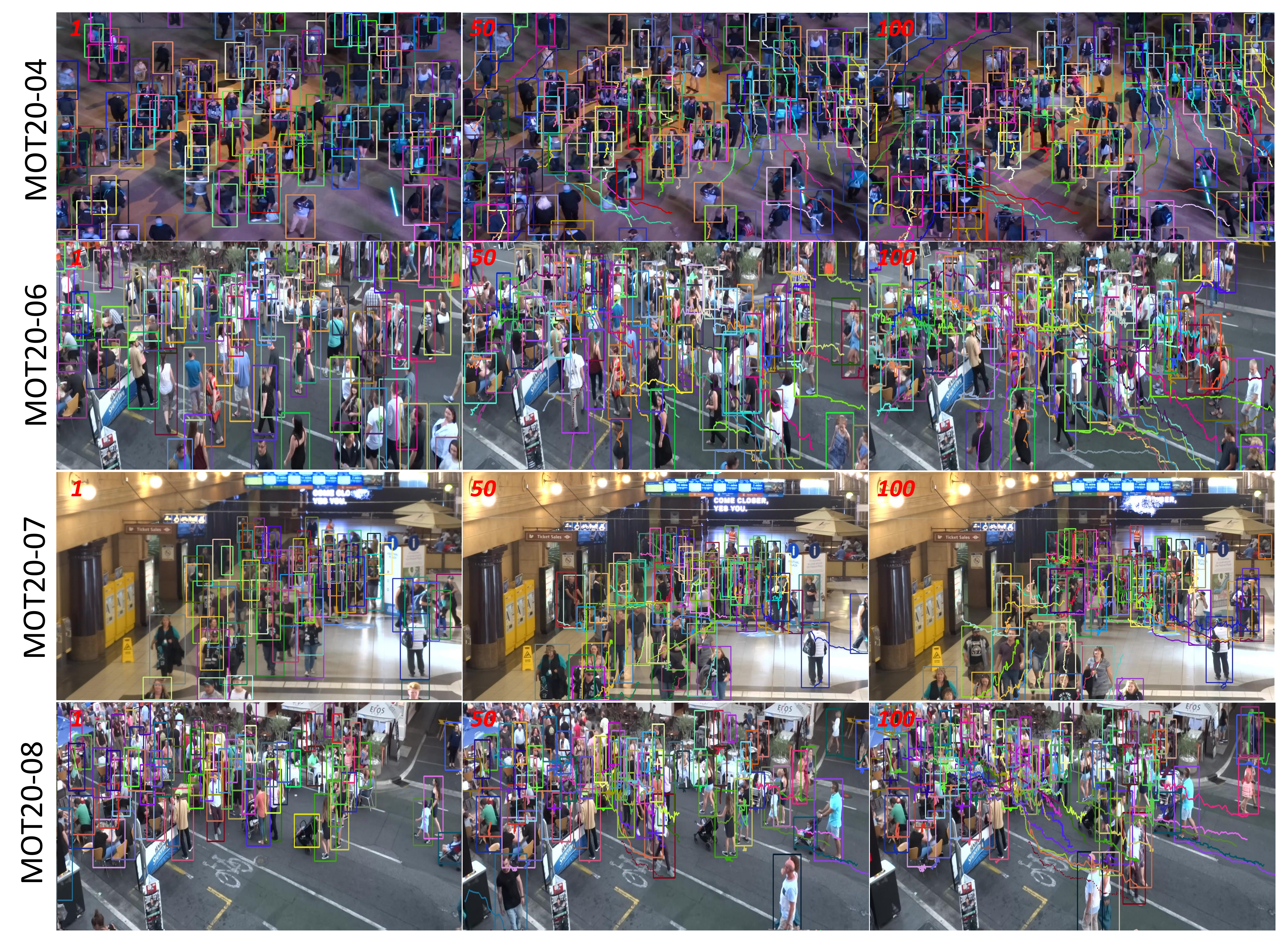}}
\vspace{0.3em}
\caption{Qualitative results of our CSTrack on MOT20~\cite{mot20}. Different colored bounding boxes denote different identity. The line under each bounding box indicates the tracklet of each target. The frame number is in the upper left of the figure. Best viewed in color and zoom in.}
\label{fig:vis_20}
\vspace{1em}
\centerline{\includegraphics[height=12\baselineskip]{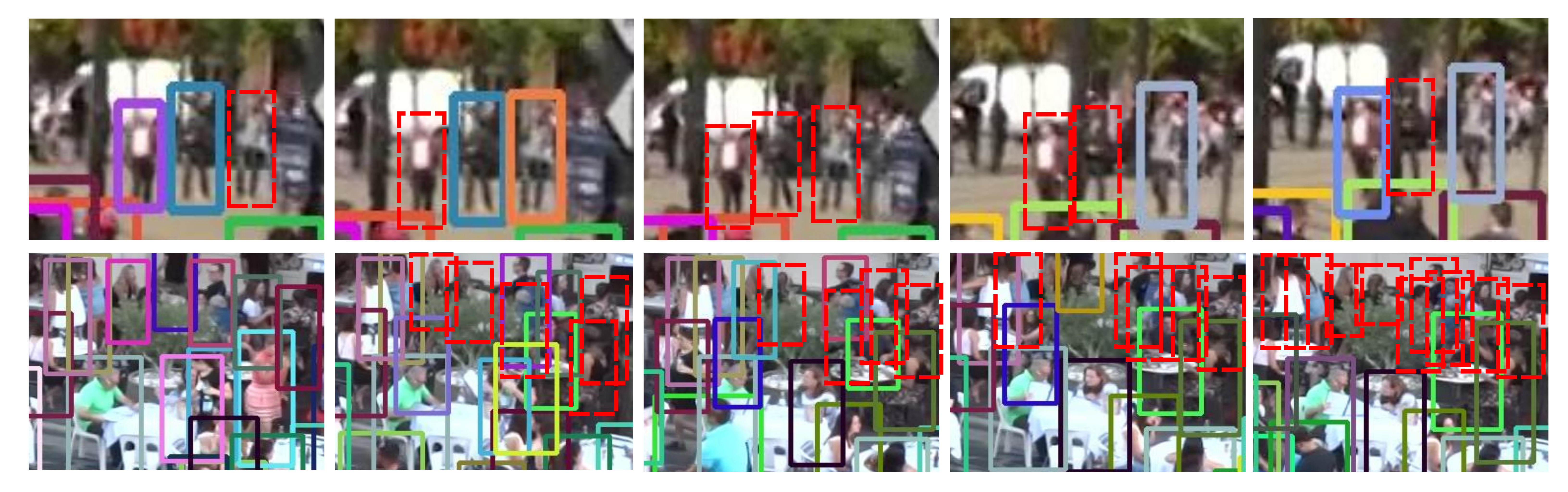}}
\caption{Failure cases: the first row (a) illustrates image blur and the second row (b) illustrates over-occlusion. The red dotted boxes indicate false negatives.}
\label{fig:failures}
\end{figure*}

\textbf{Discriminative Ability of ID Embeddings.} 
We visualize the discriminative ability of ID embeddings between JDE and our method. As shown in Fig.~\ref{fig:reid} (b) and (d), we observe that ID embeddings of JDE look more similar among targets than CSTrack, not only in the current example frame but also in the temple (\textit{i.e.,} ID embeddings of the previous tracklets).  For a more straightforward comparison, we construct the cosine metric matrix between ID embeddings of the current example frame in Fig.~\ref{fig:reid} (c) and the cosine metric matrix between the templates in Fig.~\ref{fig:reid} (e). Here, red color indicates a higher correlation between two ID embeddings, and blue indicates less correlated. The value on the diagonal indicates the similarity between the target and itself. The ideal cosine metric matrixes in Fig.~\ref{fig:reid} (c) and Fig.~\ref{fig:reid} (e) are ones with only the diagonals red and the rest blue. Compared with JDE~\cite{JDE}, we observe that our method can learn more discriminative embeddings to avoid ambiguous matching between targets. It is also efficient for the update of the discriminative template. Moreover, we visualize the cosine metric matrix between ID embeddings of the current frame and previous tracklets template in Fig.~\ref{fig:reid} (f). During matching, we hope that each row and each column has at most one red color in ideal cosine metric matrix, and the rest are blue. As shown in Fig.~\ref{fig:reid} (f), we observe that our method can significantly improve the association ability. The advantage over JDE~\cite{JDE} is most pronounced on the robustness to occlusion and scale changes (see Fig.~\ref{fig:reid} (a)), which verifies that our proposed SAAN network is efficient to alleviate scale-aware competition and improve the resilience to objects with different scales.


\textbf{Qualitative Results and Failure Cases.} 
\label{Visualization}
In this section, we show the qualitative results in MOT17 testing set~\cite{mot16} (see Fig.~\ref{fig:vis_17}) and MOT20 testing set~\cite{mot20} (see Fig.~\ref{fig:vis_20}). We observe that our method is efficient to deal with large-scale variations and keep correct identities (see MOT17-08 and MOT17-12). It mainly attributes to our SAAN module as it  focuses on multi-resolution information to learn more discriminative embeddings. Moreover, we observe that our tracker performs well in many challenging scenes. For instance, from the results of MOT17-07 and MOT17-14, we can find that our tracker can predict accurate bounding boxes for small targets. In the visualization of MOT17-03, we find that our tracker performs robustly to occlusion objects, even in the presence of over-crowding (\textit{e.g.,} the scenes in MOT20 testing set).

Despite the promising results, we also should be aware of the unaddressed cases. We discuss two scenarios in which CSTrack fails in Fig.~\ref{fig:failures}.  In particular, Fig.~\ref{fig:failures} (a) illustrates the negative influence of image blur, which makes our detector misclassify objects as background at some points. Fig.~\ref{fig:failures} (b) shows another failure circumstance, i.e., over-occlusion. As we can see, only a fraction of the targets have been detected over time, and the same local-visible objects will be missed by the detector. These cases are challenging for MOT because the missed targets will break the temporal consistency of tracklets. Despite being different in nature, these two cases arguably arise from the over-rely on image-based detection, which only focuses on single-frame features from the current frame. When a single-frame feature is not reliable, the tracker will miss the objects. In further work, we will further improve the one-shot tracker by employing temporal information.

\section{Conclusion}
\label{sec:conclusion}
In this paper, we propose a new one-shot online model CSTrack for the MOT task. A novel reciprocal network (REN) and scale-aware attention network (SAAN) are introduced to mitigate the competition and improve the collaboration of detection and ReID subtasks in MOT. The experimental results demonstrate the effectiveness and the efficiency of our framework. Compared with the methods on public benchmarks, our model achieves state-of-the-art performance, which outperforms our baseline JDE by +11.2\% on MOTA and +17.5\% on IDF1. Besides, CSTrack is a nearly real-time MOT tracker and its lightweight version can run at 34.6 FPS with only a minor performance drop, which is more suitable for real applications.

\section*{Acknowledgment}
This work was supported by the Natural Science Foundation of China (No. 61972071, U20A20184), the Sichuan Science and Technology Program (2020YJ0036), the Research Program of Zhejiang Lab (2019KDAB02), the Open Project Program of the National Laboratory of Pattern Recognition (201900014), and Grant SCITLAB-1005 of Intelligent Terminal Key Laboratory of SiChuan Province.

\bibliographystyle{IEEEtran}
\bibliography{cstrack}
%

\vspace{-3em}
\begin{IEEEbiography}[{\includegraphics[width=1in,height=1.25in,clip,keepaspectratio]{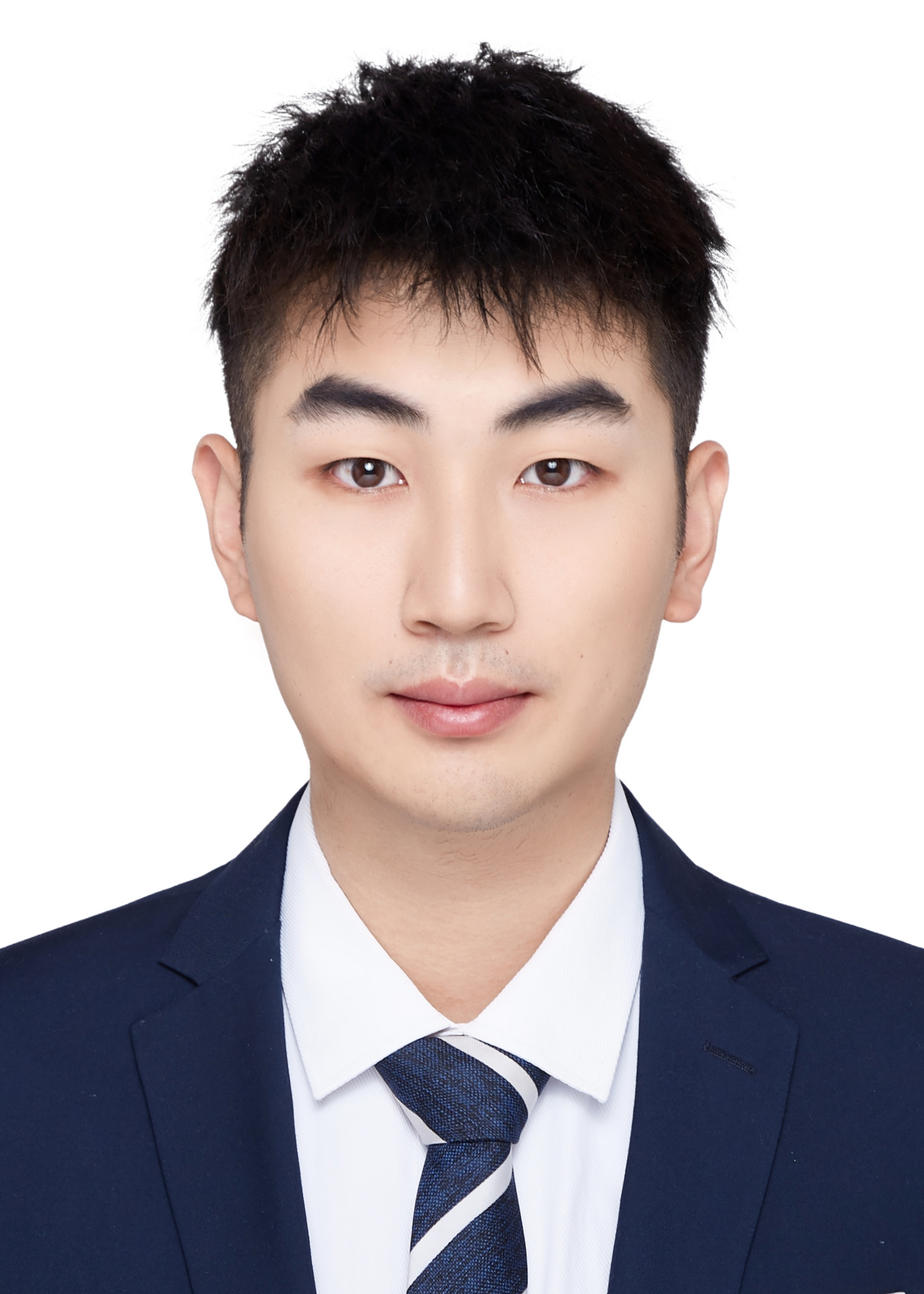}}]{Chao Liang}
received a Bachelor’s degree from School of Mechanical Engineering, Southwest Jiaotong University, Chengdu, China, in 2019. From 2019 to now, he is pursuing his M.S. degree in School of Automation Engineering, University of Electronic  Science and Technology of China, Chengdu, China.  His research interests include object detection, object tracking and person re-identification.
\end{IEEEbiography}

\vspace{-3em}
\begin{IEEEbiography}[{\includegraphics[width=1in,height=1in,clip,keepaspectratio]{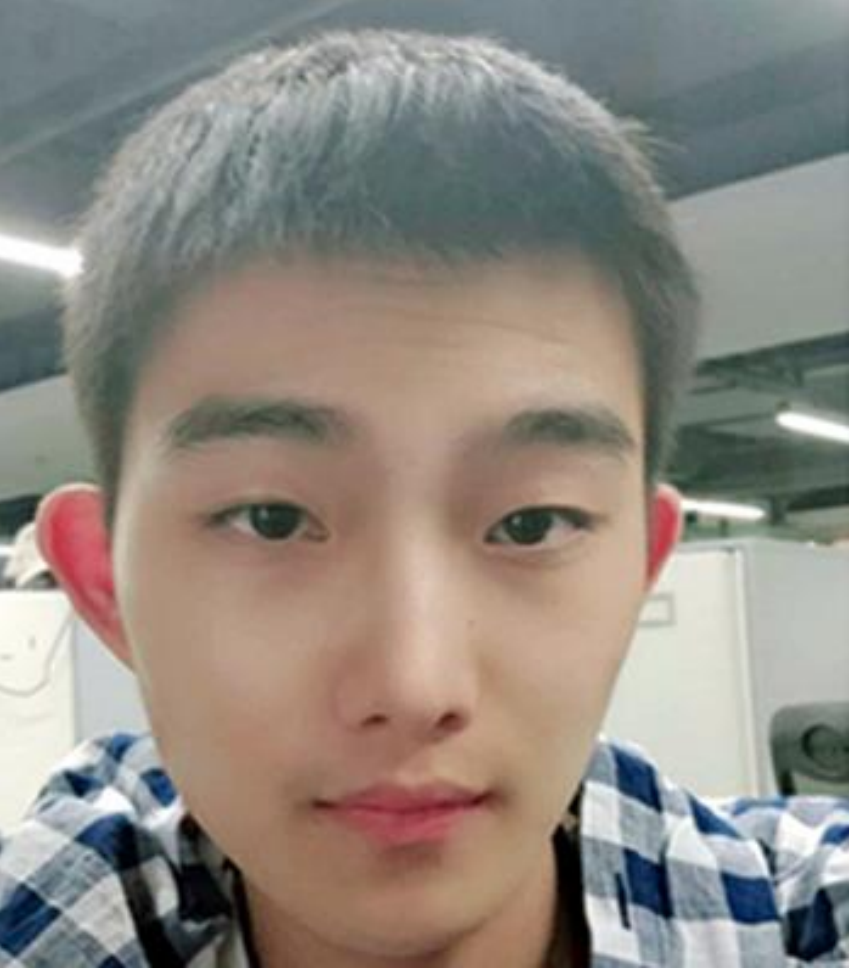}}]{Zhipeng Zhang}

received a Bachelor's degree in optics and information engineering from the University of Electronic Science and Technology of China in 2017. From 2017 to now, he is reading a Ph.D. of pattern recognition and intelligent system in the Institute of Automation, Chinese Academy of Sciences. His current research interest is video understanding, including object tracking, object segmentation, and video representation learning.

\end{IEEEbiography}

\vspace{-5em}
\begin{IEEEbiography}[{\includegraphics[width=1in,height=1.25in,clip,keepaspectratio]{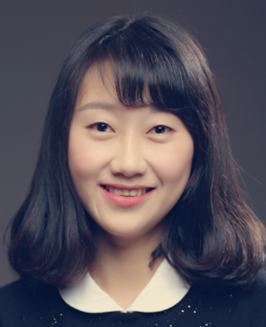}}]{Xue Zhou}
(Member, IEEE) received the B.S. degree in automatic control from the University of Electronic Science and Technology of China, Chengdu, China, in 2003, and the Ph.D. degree in pattern recognition and intelligent system from the Institute of Automation, Chinese Academy of Sciences, Beijing, China, in 2008. She is currently a Professor with the School of Automation Engineering, University of Electronic Science and Technology of China. Her
current research interests are video object perception, including video object segmentation, level-set based object tracking, multiple object tracking and pedestrian re-identification.
\end{IEEEbiography}

\vspace{-5em}
\begin{IEEEbiography}[{\includegraphics[width=1in,height=1.25in,clip,keepaspectratio]{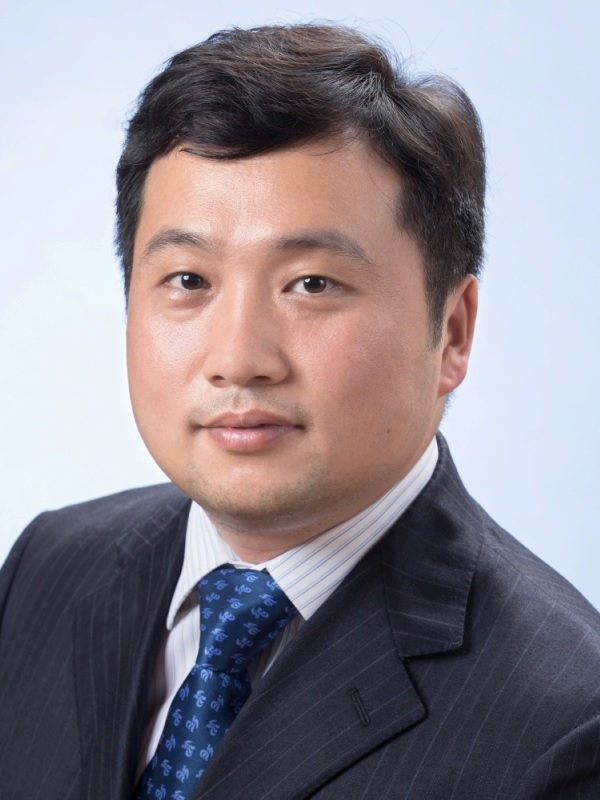}}]{Bing Li}
received the Ph.D. degree from the Department of Computer Science and Engineering, Beijing Jiaotong University, Beijing, China, in 2009. He is currently a Professor with the Institute of Automation, Chinese Academy of Sciences, Beijing. His current research interests include video understanding, color constancy, visual saliency, multi-instance learning, and Web content security.
\end{IEEEbiography}

\vspace{-4em}
\begin{IEEEbiography}[{\includegraphics[width=1in,height=1.25in,clip,keepaspectratio]{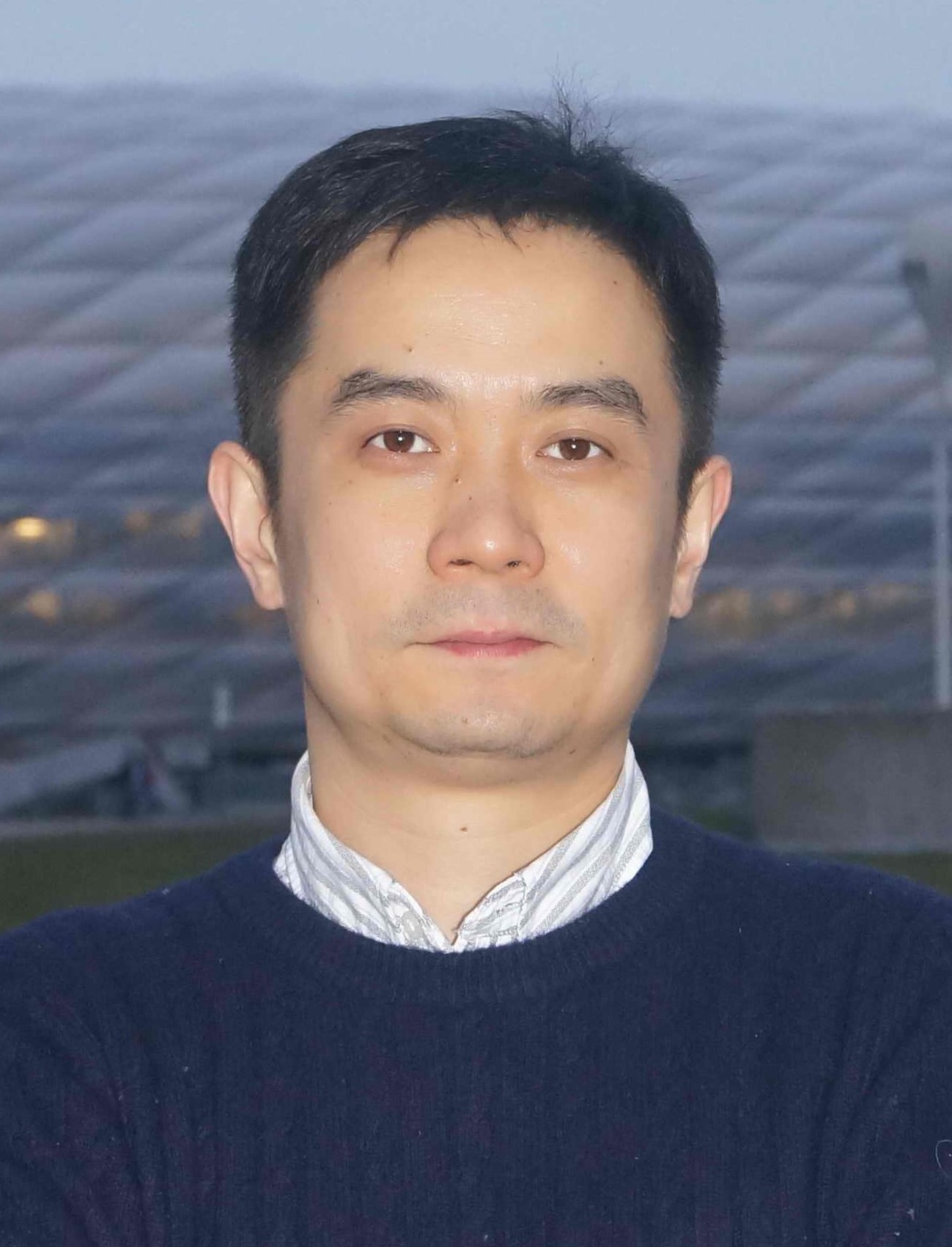}}]{Shuyuan Zhu}(S'08-A'09-M'13) received the Ph.D. degree from the Hong Kong University of Science and Technology (HKUST), Hong Kong, in 2010. From 2010 to 2012, he worked at HKUST and Hong Kong Applied Science and Technology Research Institute Company Limited, respectively. In 2013, he joined University of Electronic Science and Technology of China and is currently a Professor with School of Information and Communication Engineering. Dr. Zhu's research interests include image/video compression and image processing. He currently serves as an Associate Editor of IEEE Transactions on Circuits and Systems for Video Technology.
\end{IEEEbiography}

\vspace{-4em}
\begin{IEEEbiography}[{\includegraphics[width=1in,height=1.25in,clip,keepaspectratio]{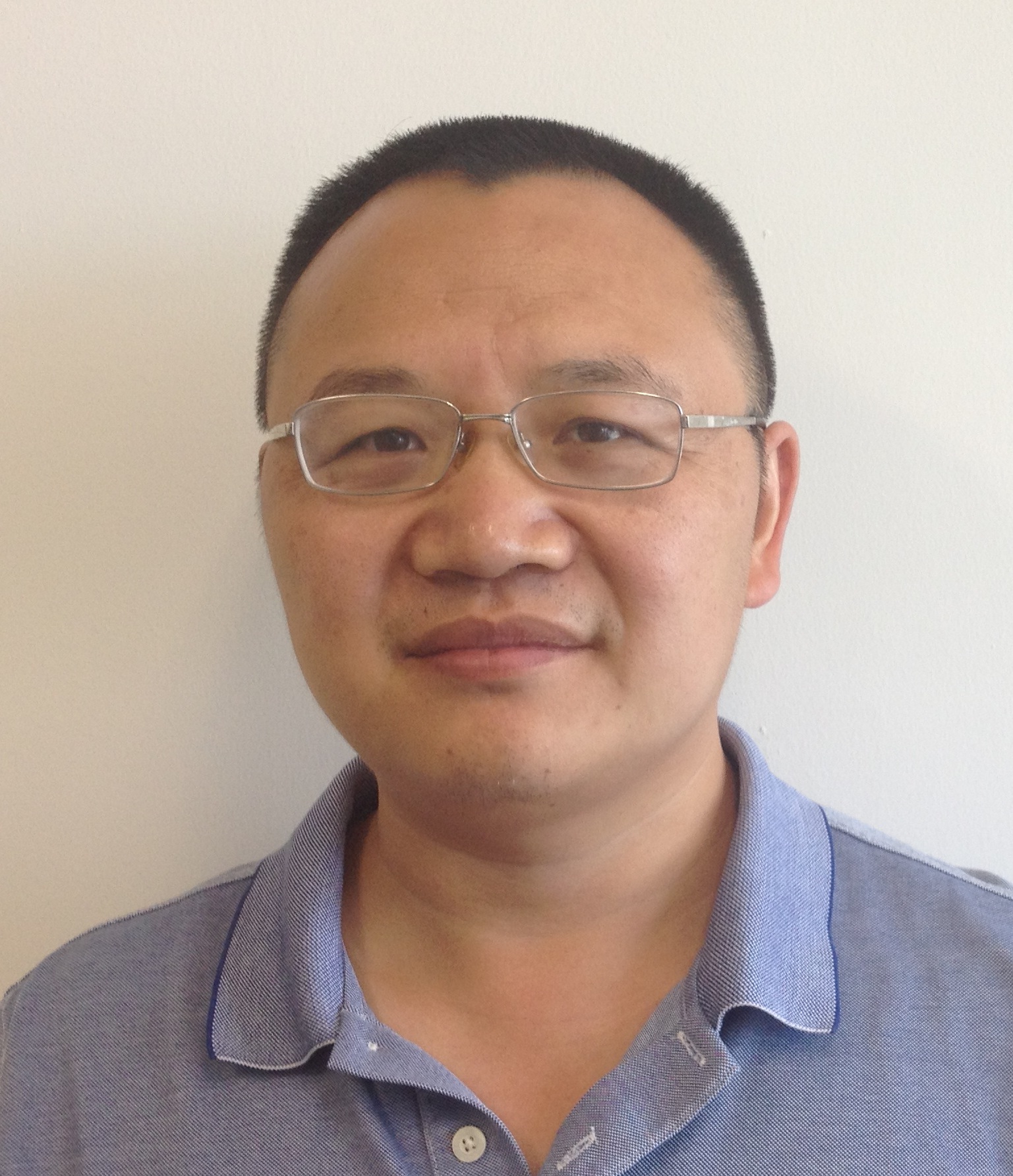}}]{Weiming Hu}
received the Ph.D. degree from the Department of Computer Science and Engineering, Zhejiang University, Zhejiang, China, in 1998. From 1998 to 2000, he was a postdoctoral research fellow with the Institute of Computer Science and Technology, Peking University, Beijing. He is currently a professor with the Institute of Automation, Chinese Academy of Sciences(CASIA), Beijing. His research interests are visual motion analysis, recognition of web objectionable information, and network intrusion detection.

\end{IEEEbiography}

\end{document}